\documentclass[11pt]{article}

\usepackage{acl}

\usepackage{times}
\usepackage{latexsym}

\usepackage[T1]{fontenc}

\usepackage[utf8]{inputenc}

\usepackage{microtype}

\usepackage{inconsolata}

\usepackage{graphicx}
\usepackage{xspace}
\usepackage{booktabs}
\usepackage{cleveref}
\usepackage[textsize=tiny]{todonotes}
\usepackage{multirow}
\usepackage{caption}
\usepackage{subcaption}
\usepackage{enumitem}
\usepackage{listings}
\newcommand{\dataname}{\textsc{IFllm}\xspace}
\newcommand{\heading}[1]{\vspace{5pt}\noindent\underline{\textsc{#1}}}

\title{Your Mouse and Eyes Secretly Leak Your Preference: \\ LLM Alignment using Implicit Feedback from Users}

\author{Haw-Shiuan Chang\thanks{\, indicates equal contribution.}\textsuperscript{1} \;\; Jeffrey Gomez\footnotemark[1]\textsuperscript{1} \;\; Mehul Patwari\footnotemark[1]\textsuperscript{1} \\ 
\textbf{Aryan Sajith}\thanks{\, The work is done at UMass Amherst.}\textsuperscript{2} \;\; \textbf{Hamed Zamani}\textsuperscript{1} \\
        \textsuperscript{1}University of Massachusetts, Amherst, USA \\
        \textsuperscript{2}York University, Canada \\
\texttt{hschang@cs.umass.edu, \{jggomez, mpatwari\}@umass.edu}, asajith@yorku.ca \\ \texttt{zamani@cs.umass.edu} \\
}

\begin{document}

\maketitle

\begin{abstract}
To align a Large Language Model (LLM), most existing methods collect explicit human feedback and train a reward model to predict the human preference based on the response text. These existing methods have two key limitations. First, the users rarely provide explicit feedback for LLM responses, which makes the high-quality preference annotation expensive to collect. Second, the methods do not leverage implicit human feedback, which has proven vital to the economic moats of Internet giants. To quantify the value of implicit feedback, we build a new dataset called \dataname, which collects $1336$ multi-turn questions from the $59$ Mechanical Turk workers, their mouse trajectories, and eye gazing points to the LLMs' responses from their webcams. \dataname shows that the users have very diverse types of gazing behavior and mouse trajectories. Our reward model based on the implicit user feedback boosts the accuracy of the text-based reward model from 55\% to 64\% and nearly triples the relative response quality improvements after applying the DPO to eight LLMs, demonstrating the value of implicit feedback in the wild. Our data collection website, dataset, and codes can be found at \url{https://github.com/themehulpatwari/llm-implicit-feedback/}. 






\end{abstract}

\begin{figure}[t!]
  \includegraphics[width=0.99\linewidth]{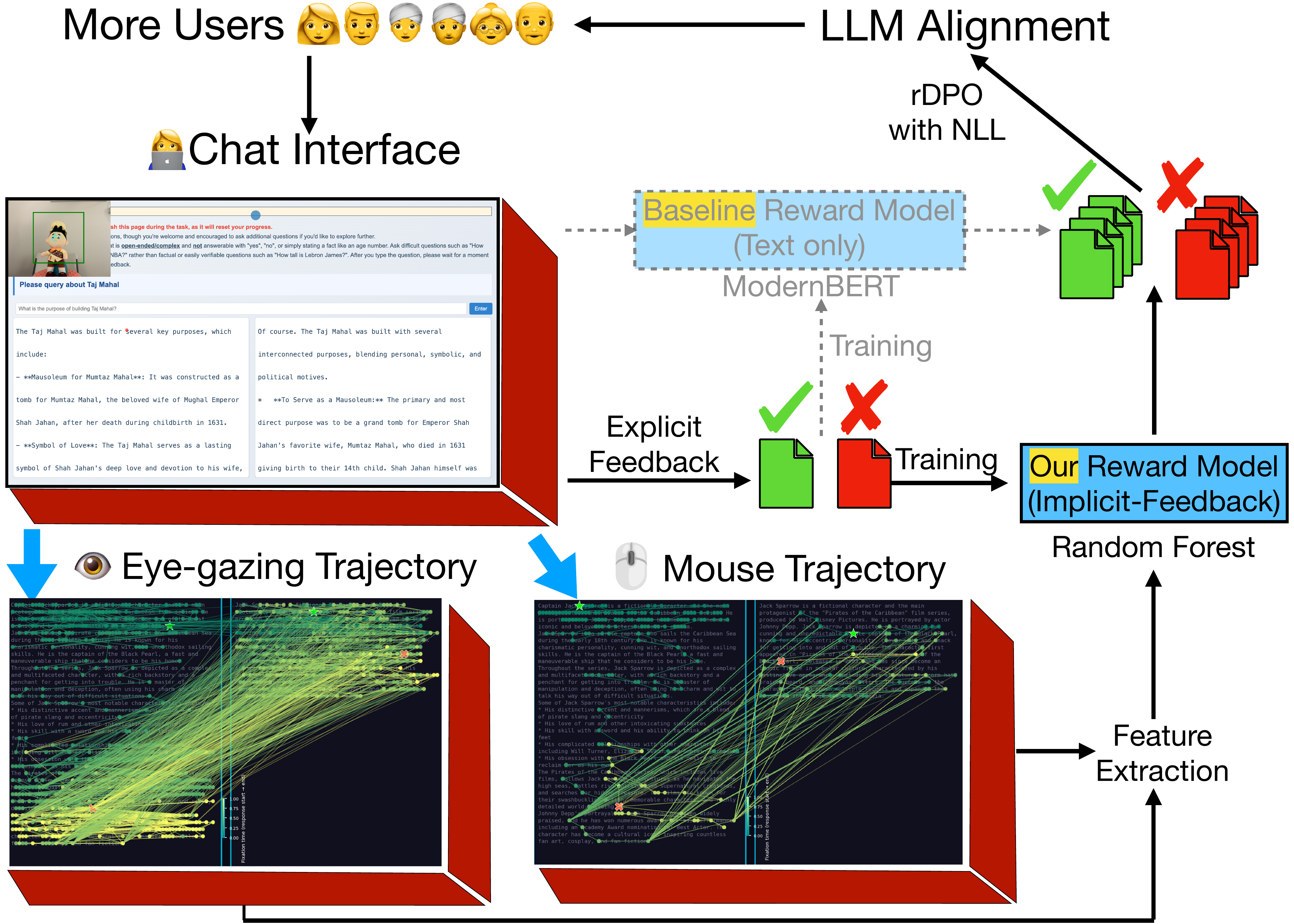} 
  \caption{ \dataname records the trajectories of eye gazing and mouse from a question answering session between a user and two LLMs. Then, we train our random forest reward model on the features extracted from the trajectories and preference labels from the user. Finally, we show that applying DPO to preferences predicted by our reward model improves LLM outputs more than a standard text-based reward model. This improvement could attract more users, enrich implicit user preferences, and promote a positive feedback loop.
  }
  \label{fig:first_fig}
\end{figure}

\section{Introduction}
\label{sec:}
Large-scale intelligent systems deployed in industry are designed to satisfy user needs and align with user expectations. In early stages of system development, researchers and practitioners typically rely on assumptions about these expectations, informed by prior experience and limited user interviews. These assumptions are subsequently operationalized through the design of annotation guidelines and the collection of labeled data used to optimize system performance. However, this paradigm does not scale effectively. Large-scale human annotation is both costly and time-intensive, and the resulting data often fails to accurately reflect real-world user interactions.

To address this limitation, some user-facing LLM providers, such as OpenAI, incorporate explicit user feedback on generated responses~\cite{han2025reinforcement}. This strategy is particularly important given that only 1–3\% of users provide feedback such as thumbs-up or thumbs-down \cite{wang2025drift}. Moreover, prior work suggests that frequent solicitation of explicit feedback can negatively impact user satisfaction~\cite{Zhao:2018}.

This is why prior successful intelligent systems, such as search engines and recommender systems, have extensively used implicit feedback signals for improving their system. For instance, click data is an important signal (if not the most important signal) in training ranking models in search engines \cite{Joachims:2002} and recommender systems \cite{oard1998implicit}. Despite tremendous success in using implicit feedback in these technologies, implicit feedback has been relatively underexplored for improving user-facing LLM technologies. A main reason is that common implicit feedback signals used in prior systems, such as clickthrough data, barely exist in many of these systems \cite{Allan:2024}. In other words, users infrequently engage with links provided by LLMs, when such links are available, and even when these interactions occur, it remains unclear how to reliably interpret them as training signals. \textit{Therefore, this paper investigates implicit feedback in the context of user-facing LLM systems and examines its potential for improving model alignment.} In this work, we focus on two forms of implicit feedback: (1) \textit{mouse movement}, which is readily available at scale in real-world deployments, and (2) \textit{eye tracking}, which, although not yet widely accessible, is representative of a broader class of multimodal user signals. We anticipate a future in which intelligent assistants can leverage such inputs, including gaze patterns, facial expressions, and hand gestures, to better model user intent and improve system alignment. To improve the practical feasibility of eye tracking signals, we rely on webcams, as opposed to special-purpose eye trackers, which are often only available in controlled lab environments. We have performed extensive efforts in developing a webcam-based crowdsourcing website that can be calibrated per user to work effectively with different cameras, internet browsers, and screen sizes and resolutions. The developed crowdsourcing website is released publicly for future use. 




Building upon our developed website, we first collect a new dataset, \dataname (\textbf{I}mplicit \textbf{F}eedback for \textbf{L}arge \textbf{L}anguage \textbf{M}odels), which 
contains $1336$ multi-turn question-answering interactions collected from $59$ unique Amazon Mechanical Turk workers across hundreds of topics from Wikipedia. During each interaction, users choose the topics they want to learn about, ask at least three questions, either score a single response (pointwise setting) or compare a pair of responses from different LLMs (pairwise setting), and answer some post-task questions. During the study, users' mouse movements and gaze trajectories are continuously recorded, with their consent. 

Using \dataname, we first analyze how users read and evaluate LLM responses. We show that user behavior is highly diverse and strongly influenced by response length. For example, mouse trajectories become increasingly correlated with user gazing trajectories for long responses because users must scroll through the generated response to read further. Our experiments show that the features extracted from the mouse trajectories are essential to our random forest preference classifier while the gazing signal is helpful when the responses are short. Through SFT and DPO, we train the LLMs to produce the responses that are more likely to be pointed by users' mouse and thus receiving higher scores from an LLM judge and human annotators.

\heading{Main Contributions}
\begin{itemize}[itemsep=2pt, parsep=0pt, topsep=1pt,leftmargin=*]
    \item We build \dataname, the first dataset that contains mouse and eye-gazing trajectories as well as explicit user preference for LLM responses in a realistic, multi-turn conversational setting. We also release our website source code for collecting \dataname under Apache 2 license to facilitates future implicit feedback data collection. 
    \item This paper provides the first systematic in-depth analysis of diverse reading behaviors and compares the effectiveness of mouse and gaze signals for training reward models.
    \item We show that implicit feedback, especially mouse movement for longer responses, substantially boosts preference prediction accuracy and raises the DPO improvements from $0.12$ to $0.35$. 
\end{itemize}

We posit that our findings, together with the released website and dataset, lay the groundwork for next-generation alignment methods that not only improve average LLM performance, but also enable scalable personalized alignment, which is known to significantly enhance user satisfaction \cite{salemi-etal-2024-lamp}.

\begin{figure*}[t!]
  \includegraphics[width=0.99\linewidth]{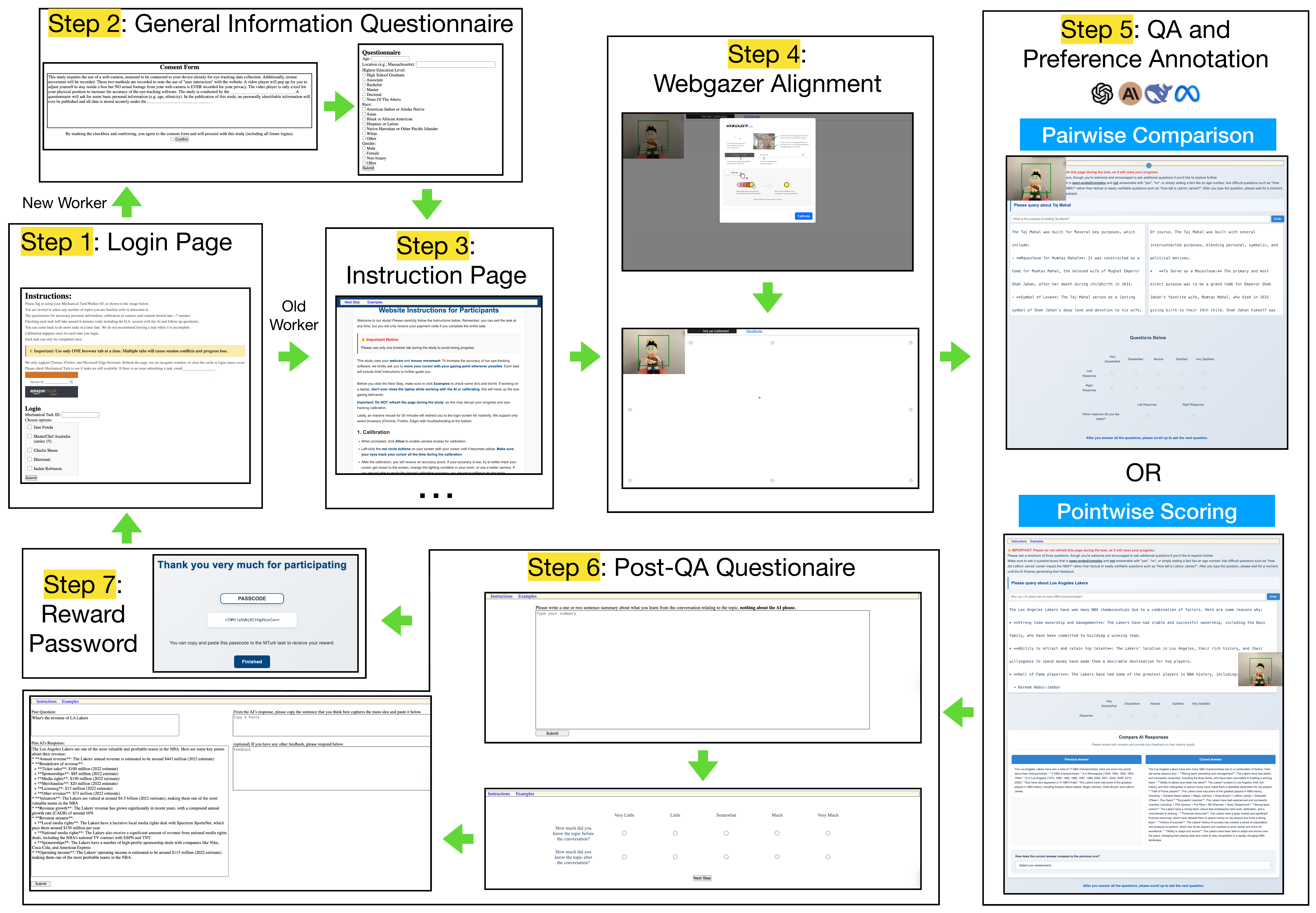} 
  \caption{Diagram of webpage navigation for a worker. 1 cycle of the webpages correlates to 1 task, equivocally 1 topic was conversed and annotated. 
  Steps 1-4 prepare the user for a task. We record the eye-gazing and mouse movement data in Step 5. After the user complete the questionaires in Steps 6, they can use the password in Step 7 to claim their reward in MTurk.  
  }
  \label{fig:website_overview}
\end{figure*} 






\section{Related Work}

Implicit feedback has been shown to be a valuable signal. Researchers have improved LLMs through human edit~\cite{gao2024aligning}, human responses in a multi-turn conversation~\cite{shi2024wildfeedback,wang2025drift}, and click and copy behavior~\cite{wang2026implicitrm}. However, they do not study the effectiveness of the eye-gazing and mouse signal.




Eye gazing data could help (large) language models in many different ways. It could be used to estimate the weights of each token in supervised fine-tuning (SFT)~\cite{zhang2025eyemulator}, align the attention of the transformer with human attention~\cite{zhang2024eyetrans,yan2024voila}, rearrange the order of aggregating contextualized embeddings~\cite{deng2024fine}, guide LLMs to generate text with different readability~\cite{sauberli2026controlling}, study how humans collaborate with LLMs~\cite{tang2024developer, tang2024codegrits}, and improve the reward model and LLM alignment~\cite{lopez2025seeing,papadopoulos2025eye}. However, no study collects and compares the implicit user feedback signals on LLMs' responses in the wild.  

Eye-gazing data has many applications in natural language processing~\cite{mathias2020survey} and machine learning. For example, eye-gazing data can be used to predict the linguistic acceptability~\cite{bondar2025colagaze,bondar2025aleyegnment}, predict the image preference of humans~\cite{papadopoulos2026gaze}, and analyze the interaction of humans with coding agents~\cite{yang2025rlhf, wang2025user}. However, these works do not focus on humans' implicit feedback to the LLMs' answers. One notable exception is the OASST-ETC dataset~\cite{lopez2025oasst}, which collects clean eye-gazing data in a controlled laboratory setting. Nevertheless, their reliance on special eye-tracking equipment and the neglect of valuable mouse movement data make them unsuitable for investigating whether LLMs could benefit from the usage of the general public.








\section{The Data Collection Website}


We develop a website 
for users to converse with LLMs. Users are recruited from Amazon Mechanical Turk (MTurk) under an approved Institutional Review Board (IRB) protocol. Our website allows a MTurk worker to do our tasks using multiple windows/tabs in Google Chrome, Firefox, or Microsoft.
Steps 1-7 of Figure~\ref{fig:website_overview} represent one run through of a task with one selected topic. 


\subsection{Login and Personal Questionnaire}

Step 1 is the Login Page where the worker selects the topic(s) they want to know more from a pool of 30 or 60 topics and we shuffle the topic order to avoid positional bias. 
Each topic is a Wikipedia page title chosen from the bottom of top 1000 popular search results from the Wikimedia API between 1/2023 and 5/2023. Our strategy aims to find the topics that the users have heard of but are not very familiar with. 

An input field requires the MTurk Worker ID. A new user will get redirected to Step 2: General Information Questionnaire, which asks the user to consent our data collection and provide some demographic information, or Step 3: Instruction Page for returning users.
Throughout this experiment, we ensure instructions are accessible and clear. 
For each session, the user must calibrate at Step 4: Webgazer Alignment. Webgazer.js~\cite{papoutsaki2015scalable} tracks your eye movement from the webcam and predicts your gazing points using a regression model. We use the calibration tool of Webgazer to train the eye-gazing model. The Webgazer displays the user's camera and instructs the user to position his/her head inside a green box for a better tracking accuracy.



\subsection{QA and Preference Annotation}

 Step 5 carries out the LLM conversation through the QA and Preference Annotation pages. Each topic is assigned as a pointwise scoring or pairwise comparison task. The user is instructed to ask non-factual questions to know more about the selected topic. Pairwise comparison uses two textboxes side-by-side. Pointwise scoring only uses one textbox. Each LLM response box is a random choice from DeepSeek V3, GPT-4o Mini, Claude Sonnet 4.5 (originally 3.5 but deprecated), or Llama 3.3 70B~\cite{grattafiori2024llama} with no duplicates in the pairwise setting. The LLMs were chosen for their popularity, diversity, significance, and/or being open-sourced. We also want to check if the structure of the LLM response affects the users' gazing pattern, so we randomly instruct the LLM to reply using bullet points. 
 
 The user is asked to query at least three times and spend at least $90$ seconds in this step. LLMs remain the same for each topic across all queries and can access the prior queries and answers, which allows the user to ask follow-up questions or conduct multi-turn interactions. If an LLM response is too long, it overflows the textbox with scrolling in the textbox enabled. At approximately every $0.1$ seconds, we record the character index and coordinates of gaze and mouse positions. Under the LLM response(s), there are question(s) for the $5$-point Likert quality scale and preference annotations of either they prefer the previous LLM response compared to the current in the pointwise setting or which response is preferred in the pairwise setting. The worker must finish all the annotations before asking the next question. The size of the textbox, font, and line spacing are large for user readability and better eye-gazing accuracy.




\subsection{Post-Test Questionnaires}
Step 6 is a Post-QA Questionnaire over three pages. The first page asks the user for a brief summary of the conversation for quality control. The second page questions the user with Likert scales (1-5) on the user's knowledge of the task before and after to quantify the quality of the LLM conversation. The last page provides the user opportunity to give feedback while asking to copy a sentence to test if a user gazes at sentences they deem significant. Step 7 gives a password to submit in Mechanical Turk as the final verification step that the user completed a task. If the worker choose multiple topics in Step 1, the user would directly go to Step 5 for the next topic after completing Step 7, which avoids wasting time on constantly gazing calibration.

\subsection{Quality Control}
\label{subsec:quality_control}


The first step for quality control is a minimum accuracy threshold of 70\%, a tradeoff between data size and quality from various camera specs. We only allow MTurk master workers to do the task at the beginning and to increase the diversity of workers, we accept the workers who have a $97$\% HIT acceptance rate and at least $10,000$ approved tasks. We manually checked summaries from Step 6 for quality assurance while filtering further based on empirically determined thresholds of how much eye gazing data was within the LLM response textbox(es) and the ratio of characters the users actually viewed. 

Overall, $83$ workers picked $275$ topics out of $300$ topics and complete $641$ pointwise tasks and $695$ pairwise tasks. $80\%$ of the tasks are completed by $27$ users. $39$ workers were identified as being below either of the thresholds (see Figure~\ref{fig:avg_norm_character_vs_norm_max_character} in appendix). Further manual analysis of weak or non-committal summaries leads to the removal of $24$ workers and $9.4$\% of tasks from the data collection.













\section{User Behavior Analyses}

We analyze how users read LLM responses using the gaze and mouse trajectories in \dataname. In the pairwise setting, users see two responses side by side, which we refer to as the left and right response; in the pointwise setting, they see a single response. Throughout, we report behavior over normalized time, a rescaling of each session's timestamps to $[0, 1]$ using linear interpolation, which allows sessions of different absolute duration to be compared on a common axis.

\subsection{Aggregate Reading Patterns}\label{sec:aggregate_pattern}

\begin{figure}[tbp]
  \includegraphics[width=0.99\linewidth]{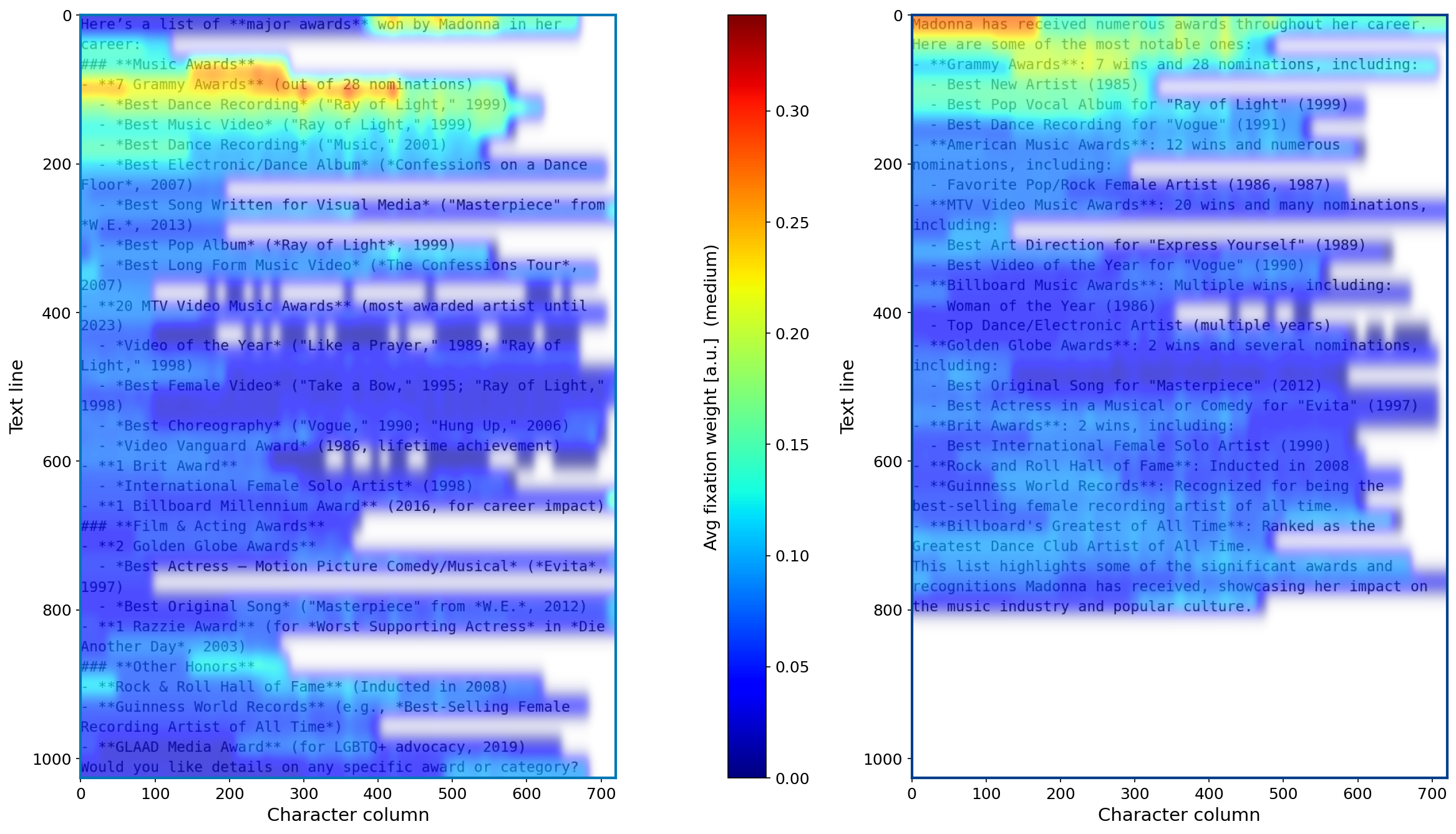}
  \caption{Average fixation weight over the response text in the pairwise setting, aggregated across all medium-length responses. The displayed text is a randomly selected example.}
  \label{fig:heatmap}
\end{figure}

\begin{figure*}[t!]
\centering
\begin{minipage}{.3\textwidth}
  \centering
  \includegraphics[width=.99\linewidth]{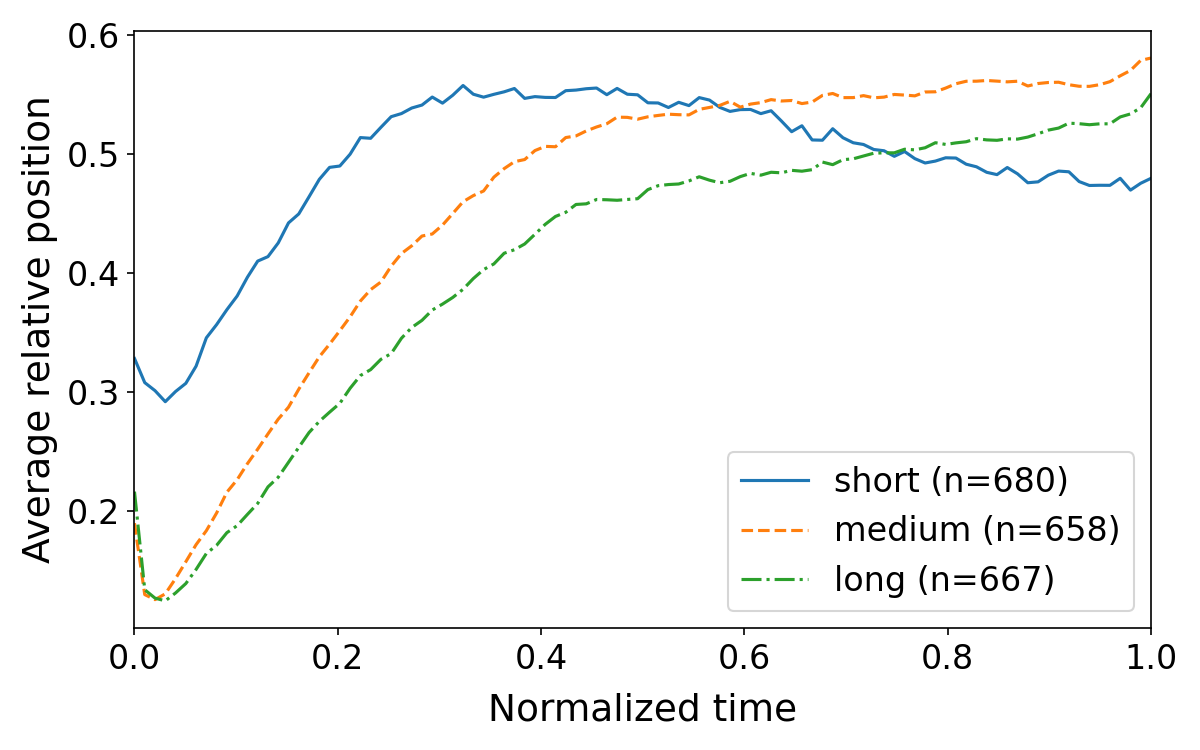}
  \captionof{figure}{Average relative gaze position over normalized time, grouped by response length (short, medium, and long responses).}
  \label{fig:length_category}
\end{minipage}%
\hspace{0.03\textwidth} 
\begin{minipage}{.3\textwidth}
  \centering
  \includegraphics[width=.99\linewidth]{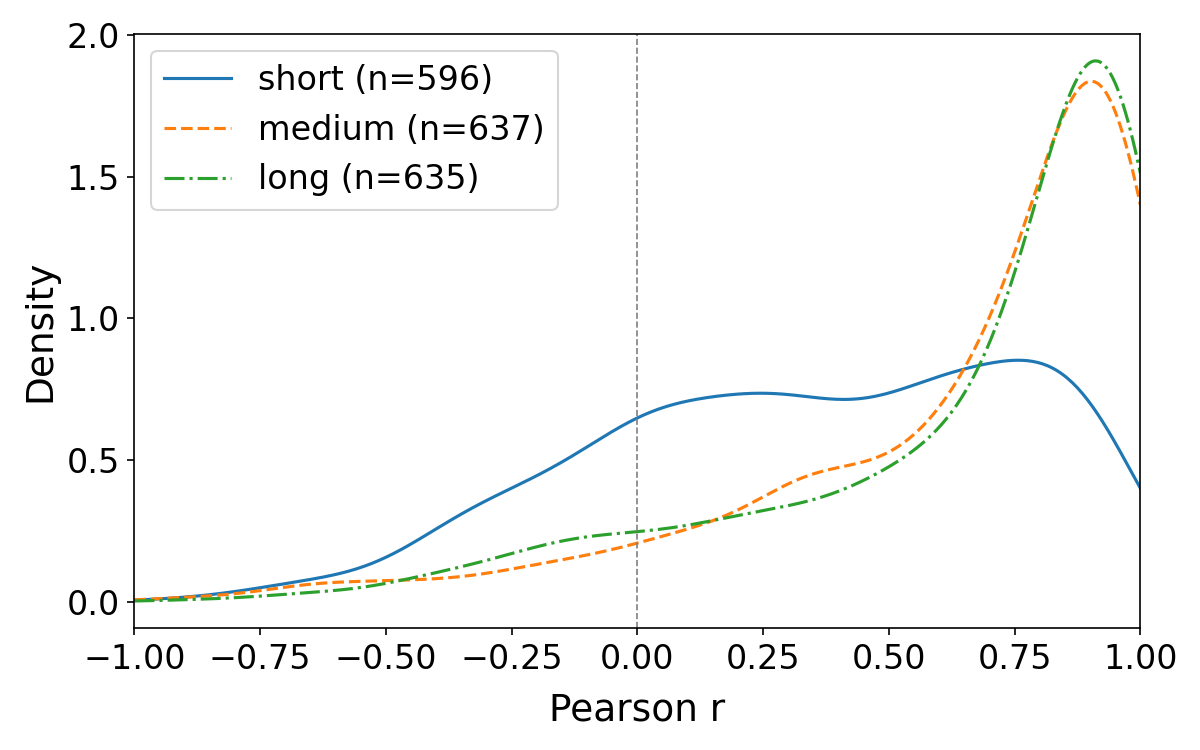}
  \captionof{figure}{Distribution of the per-session Pearson correlation between mouse and gaze position, grouped by response length.}
  \label{fig:gaze_mouse_corr}
\end{minipage}
\hspace{0.03\textwidth} 
\begin{minipage}{.3\textwidth}
\centering
  \includegraphics[width=.99\linewidth]{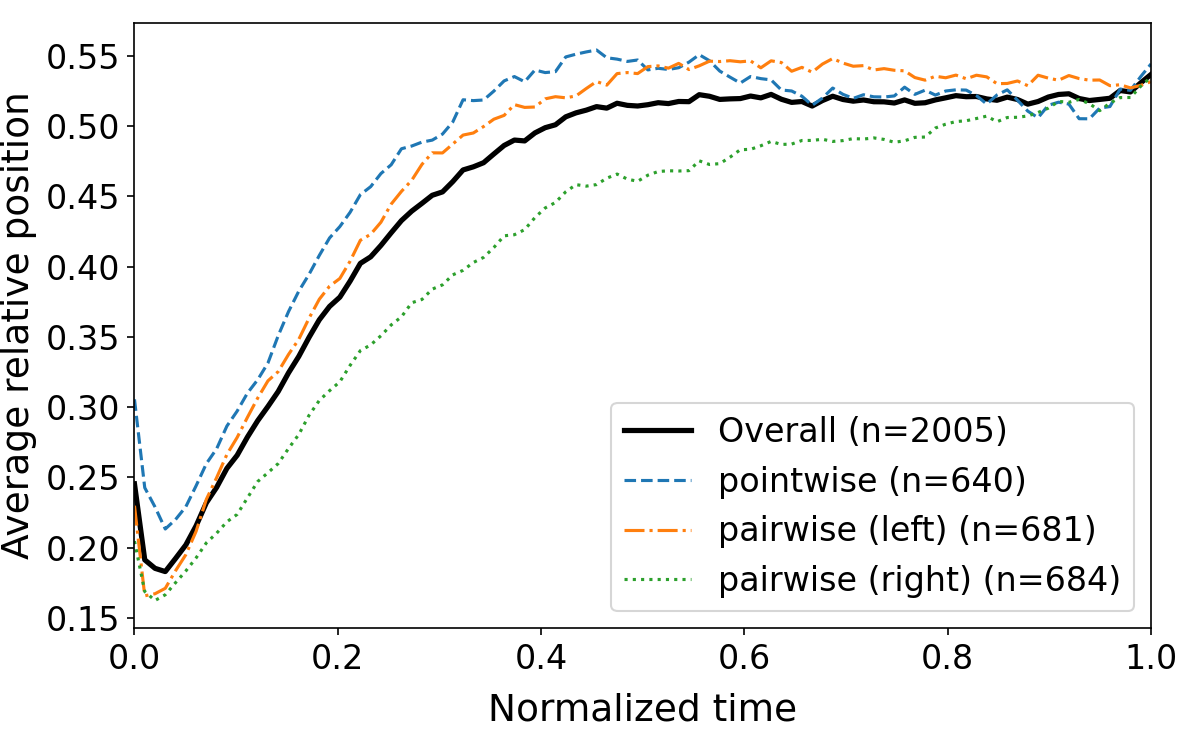}
  \captionof{figure}{Comparison of average gaze trajectories from pointwise setting and left and right responses in the pairwise setting.}
  \label{fig:pointwise_pairwise}
\end{minipage}
\end{figure*}

\begin{figure*}[t!]
\centering
\begin{minipage}{.47\textwidth}
  \centering
  \includegraphics[width=.99\linewidth]{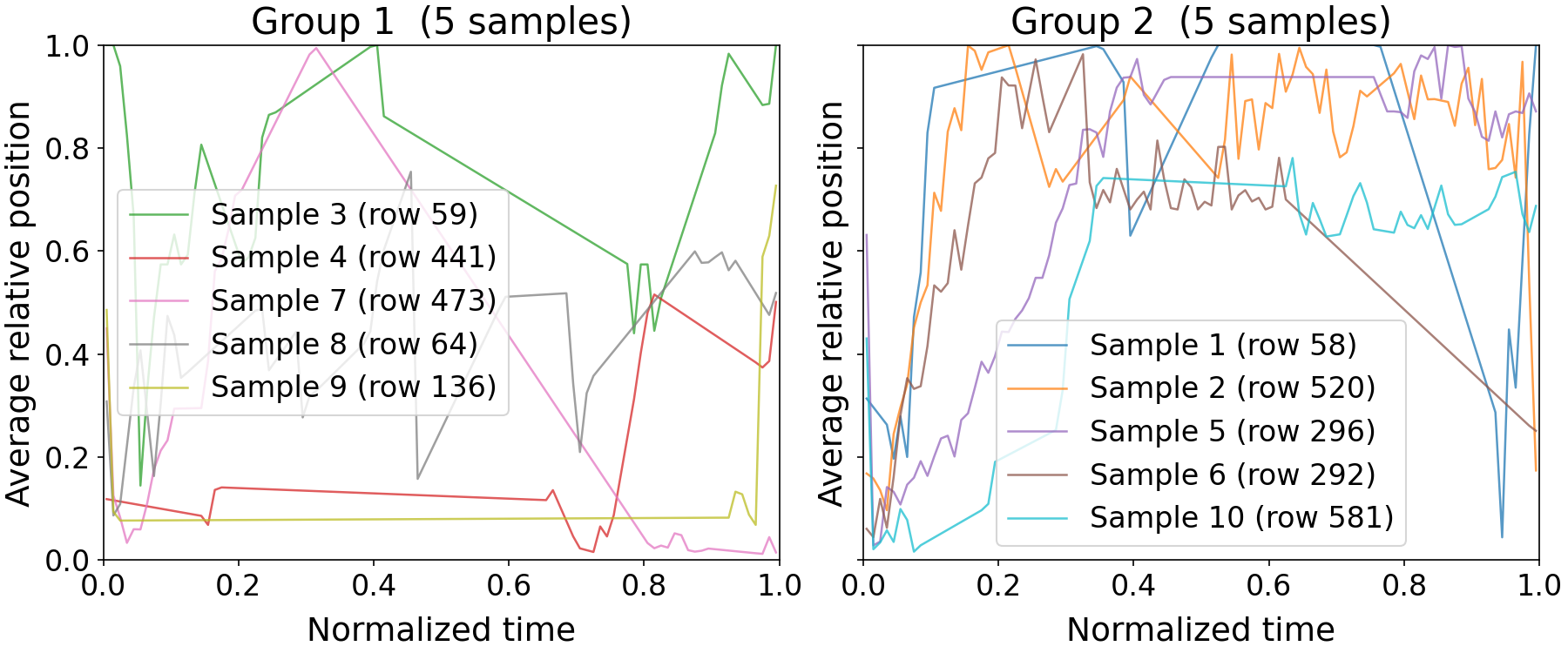}
  \captionof{figure}{Gaze trajectories of ten randomly sampled sessions over normalized time.}
  \label{fig:gaze_samples}
\end{minipage}%
\hspace{0.03\textwidth} 
\begin{minipage}{.47\textwidth}
  \centering
  \includegraphics[width=.99\linewidth]{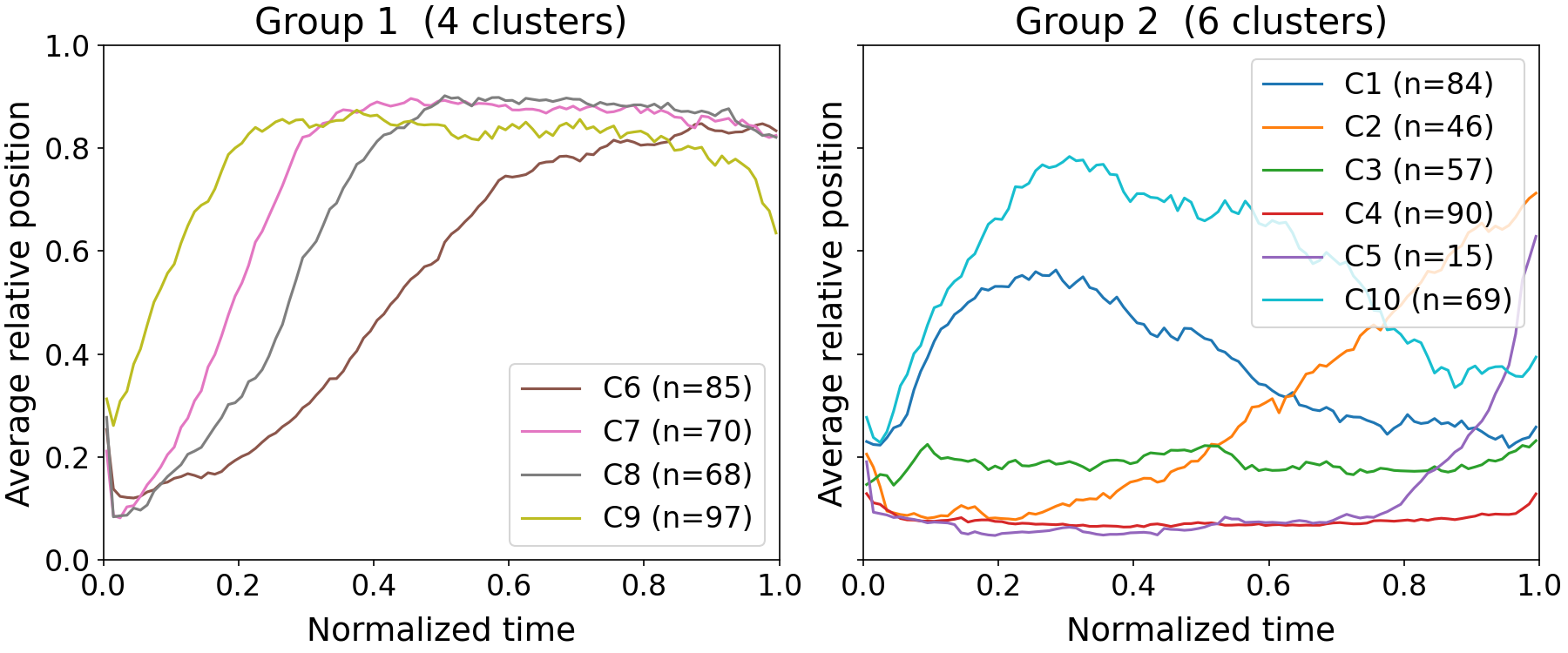}
  \captionof{figure}{Gaze trajectory clusters over normalized time. The similar cluster centers are shown in the left figure (group 1).} 
  \label{fig:gaze_clusters}
\end{minipage}
\end{figure*}

On average, users give more attention to the early part of a response than to the rest. As shown in Figure~\ref{fig:heatmap}, on the right response users concentrate on the opening words; on the left, attention shifts to the end of the first line and the start of the following lines, settling on the top-middle rather than the opening words that are commonly assumed to matter most.
The same content receives attention in different places depending on its position, which suggests that the layout of the interface changes where users direct their attention.


Figure~\ref{fig:length_category} indicates the average reading trajectory depends heavily on response length. Given a short response, users reach the end quickly and revisit text they have already read. As the response grows longer, they instead spend more time on the early portion of the response and progress more slowly.
The same length split also governs how closely the mouse follows the gaze. For short responses, Figure~\ref{fig:gaze_mouse_corr} shows that the two are only weakly correlated. For medium and long responses, they are strongly correlated, as the user must move the mouse to the text box to scroll the longer response.


Figure~\ref{fig:pointwise_pairwise} suggests that reading speed further depends on the task layout and the position of the response. Users read through the left response faster than the right. With only one response to read in the pointwise setting, users finish it early and have time to return to parts they have already seen.

\subsection{Individual Variability}


The aggregate patterns above describe the average user, but individual trajectories are highly irregular. For example, randomly selected trajectories in Figure~\ref{fig:gaze_samples} are full of back-and-forth movement, and they differ sharply from one another. 

To understand the different types of patterns, we cluster gaze trajectories with BisectingKMeans~\cite{steinbach2000comparison} and visualize the centers in Figure~\ref{fig:gaze_clusters}.
The clusters in Group 1 correspond to users who read the response and pausing to annotate or type the next query, though 
some reading quickly and others slowly. Group 2 captures the remaining styles: some users read only up to a point and then move back to what they have seen, some read at a steady rate through to the end, and some barely read the response at all.


\section{Preference Prediction}
\label{sec:pref_pred}
To simplify our description and analysis, we focus on the implicit feedback collected for the side-by-side pairwise response comparison and mention the pointwise setting, which predicts the preference between current and previous responses, as an extension. We will first extract features from the implicit feedback and train a reward model to predict users' preference.


\begin{table*}[t]
\centering
\scalebox{0.73}{
\begin{tabular}{@{}lll@{}}
\toprule
\textbf{Group} & \textbf{Feature} & \textbf{Description} \\
\midrule
\multirow{2}{*}{Text}
  & Query Length & Number of query characters \\
  & Left/Right Response Length & Character count of each response \\
\midrule
\multirow{11}{*}{Gaze}
  & Left/Right Max Character & Maximum character index read; serves as effective response length \\
  & Left/Right Norm.\ Max Character & Max Character divided by Response Length \\
  & Left/Right Total Records & Total number of gazing records ($\approx$ total seconds divided by 10) \\
  & Left/Right Total/Reviewing Points & Gazing response points before/after excluding between-review periods \\
  & Left/Right Total/Reviewing Norm. Points & Number of total/reviewing points divided by total records \\
  & Left/Right Reviewing Time & Gazing response time during review \\
  & Left/Right Reviewing Norm. Time & Reviewing gazing response time divided by total reviewing time \\
  & Left/Right Avg/Var Norm.\ Character & Mean or variance of gaze character divided by Response Length \\
  & Left/Right Avg Character in a Window & Mean gaze character in each of 20 equal time windows \\
  & Proper Head Position Ratio & Fraction of time user's head is in the WebGazer-suggested green box \\
  & Max Character Pairwise Comparison & $+1$ if left Max Character $>$ right, $-1$ otherwise  \\
  & Reviewing (Norm.)\ Time Diff & Left $-$ right reviewing (norm.) time  \\
  & Reviewing (Norm.)\ Time Ratio & Left divided by right reviewing (norm.) time \\
\midrule
Mouse & \multicolumn{2}{c}{Features identical to gaze features except for the head position feature} \\
\midrule
\multirow{1}{*}{Gaze and Mouse}
  & Left/Right Ratio of Gaze and Mouse  & Per-side Reviewing Time (Gaze) divided by Reviewing Time (Mouse) \\
\bottomrule
\end{tabular}
}
\caption{Feature descriptions for the reward models. Many gazing features have two versions. \textit{Total *}: over all records and \textit{Reviewing *} excluding estimated periods when the user annotates preferences or types a new query.}
\label{tab:features}
\end{table*}

\subsection{Feature Extraction}
\label{sec:feature_extraction}
For every 0.1 second, we record their mouse position and gaze position from WebGazer. If their gazing point is inside a text box with a response, we record the character index they gaze at and the corresponding time. We then extract features from these trajectories to train our reward model. Table~\ref{tab:features} summarizes all features; Mouse and gaze trajectories share the same file format, so mouse features mirror gaze features unless noted otherwise. In the pointwise setting, the left/right features become the current/previous features. 

\heading{Text Features:} 
Basic properties of the query and responses, including query length and response length. \citet{singhal2024long,dubois2024length} show longer responses often receive higher scores.

\heading{Gaze/Mouse Features:} When the users like a response, they tend to spend more time reading it in full~\cite{yang2025rlhf}, so we summarize the trajectories into features of reading time and position. 
A one-second smoothing window is applied to time-based features to reduce noise.


\subsection{Unused Features}
After adding the implicit feedback features above, we find that the following features are either unused by the random forest or degrade its performance, so we exclude them from our final model: LLM identity (one-hot indicators for the response source), bullet point prompt (whether the prompt included a bullet point instruction), and user identity (one-hot features for the top five most active users).

\begin{table}[t]
\centering
\scalebox{0.6}{
\begin{tabular}{lccc}
\toprule
\textbf{Method} & \textbf{Accuracy} & \textbf{F1 Class 1} & \textbf{F1 Class 0} \\
\midrule
\multicolumn{4}{c}{ \textbf{Pairwise (695 Samples)}} \\
\midrule
Claude-S-4-6 + Zero-Shot &	0.5303 &	0.5163 &	0.5434 \\
Gemma-4-31B + Zero-Shot &	0.5499 & 	0.5252 &	0.5722\\
\midrule
mBERT base + Text                 & 0.5549$_{\pm0.0347}$ & 0.5220$_{\pm0.0366}$ & 0.5815$_{\pm0.0353}$ \\
mBERT base + Text + IF  & 0.5896$_{\pm0.0193}$ & 0.4733$_{\pm0.1194}$ & 0.6319$_{\pm0.0172}$ \\
mBERT large + Text               & 0.5577$_{\pm0.0168}$ & 0.5308$_{\pm0.0194}$ & 0.5810$_{\pm0.0164}$ \\
Qwen3 1.7B + Text                     & 0.5548$_{\pm0.0261}$ & 0.5101$_{\pm0.0338}$ & 0.5906$_{\pm0.0231}$ \\
\midrule
RF + All Features                 & 0.6074$_{\pm0.0073}$ & 0.5776$_{\pm0.0132}$ & 0.6322$_{\pm0.0070}$ \\
RF + (IF - Gaze)                        & 0.6267$_{\pm0.0097}$ & 0.5935$_{\pm0.0173}$ & 0.6548$_{\pm0.0121}$ \\
RF + (IF - Mouse)                         & 0.6089$_{\pm0.0130}$ & 0.5836$_{\pm0.0162}$ & 0.6313$_{\pm0.0157}$ \\
RF + IF                              & \textbf{0.6415}$_{\pm0.0097}$ & \textbf{0.6084}$_{\pm0.0173}$ & \textbf{0.6694}$_{\pm0.0121}$ \\
\midrule
\multicolumn{4}{c}{ \textbf{Pointwise (248 Samples)}} \\
\midrule
mBERT base + Text       & 0.5239$_{\pm0.0667}$ & 0.4716$_{\pm0.1042}$ & 0.5636$_{\pm0.0495}$ \\
mBERT base + Text + IF  &  0.5319$_{\pm0.0434}$ & 0.3325$_{\pm0.1903}$ & 0.6254$_{\pm0.0648}$ \\
\midrule
RF + (IF - Gaze)           & 0.6088$_{\pm0.0379}$ & 0.6240$_{\pm0.0432}$ & 0.5846$_{\pm0.0406}$ \\
RF + (IF - Mouse)          & 0.6006$_{\pm0.0185}$ & 0.5935$_{\pm0.0252}$ & 0.6044$_{\pm0.0183}$ \\
RF + IF                    & \textbf{0.6291}$_{\pm0.0262}$ & \textbf{0.6377}$_{\pm0.0301}$ & \textbf{0.6137}$_{\pm0.0297}$ \\

\bottomrule
\end{tabular}
}
\caption{Reward model performance comparison (average $\pm$ standard error across folds). IF refers to the important features from the implicit feedbacks. RF refers to random forest. Claude-S-4-6 means Claude Sonnet 4.6. mBERT means ModernBERT.}
\label{tab:reward_results}
\end{table}

\begin{figure*}[t!]
\centering
\begin{minipage}{.43\textwidth}
  \centering
  \includegraphics[width=.99\linewidth]{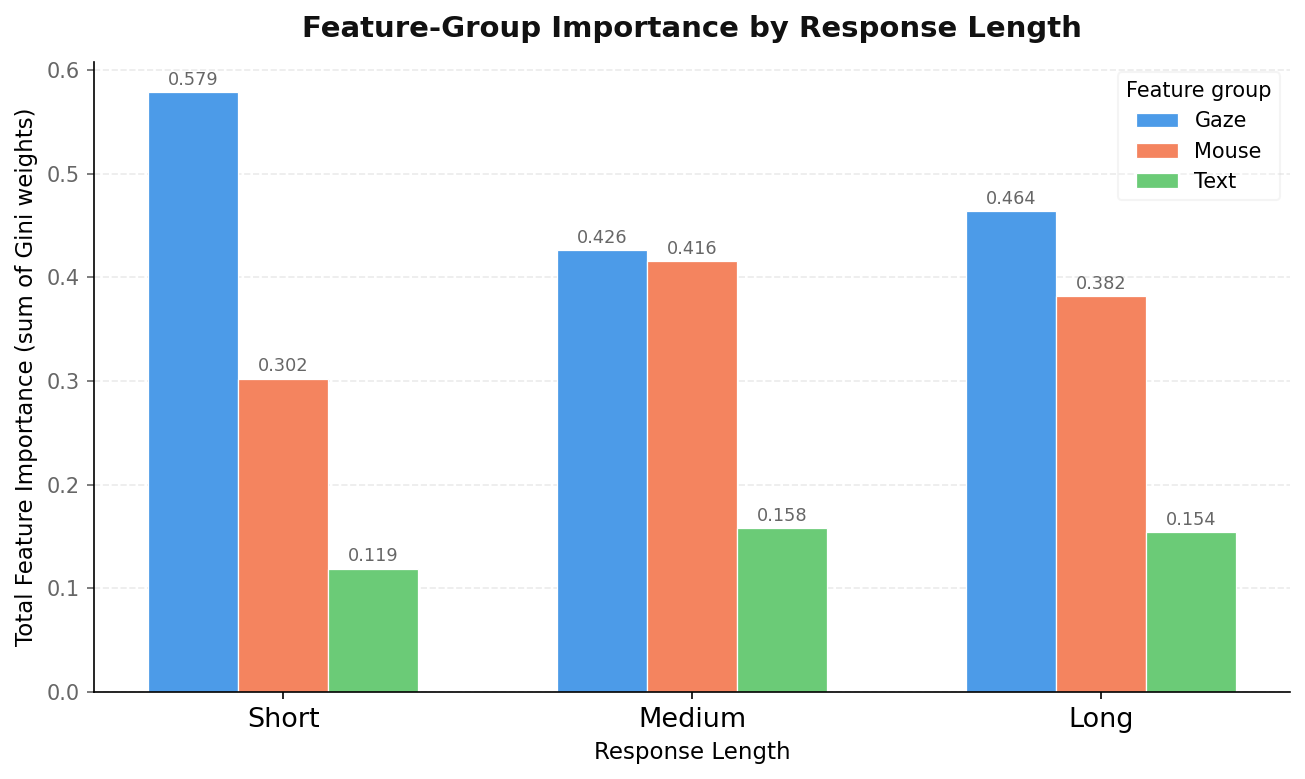}
  \captionof{figure}{The comparison of features weights given different response lengths}
  \label{fig:feature_imp_length}
\end{minipage}%
\hspace{0.03\textwidth} 
\begin{minipage}{.5\textwidth}
  \centering
  \includegraphics[width=.99\linewidth]{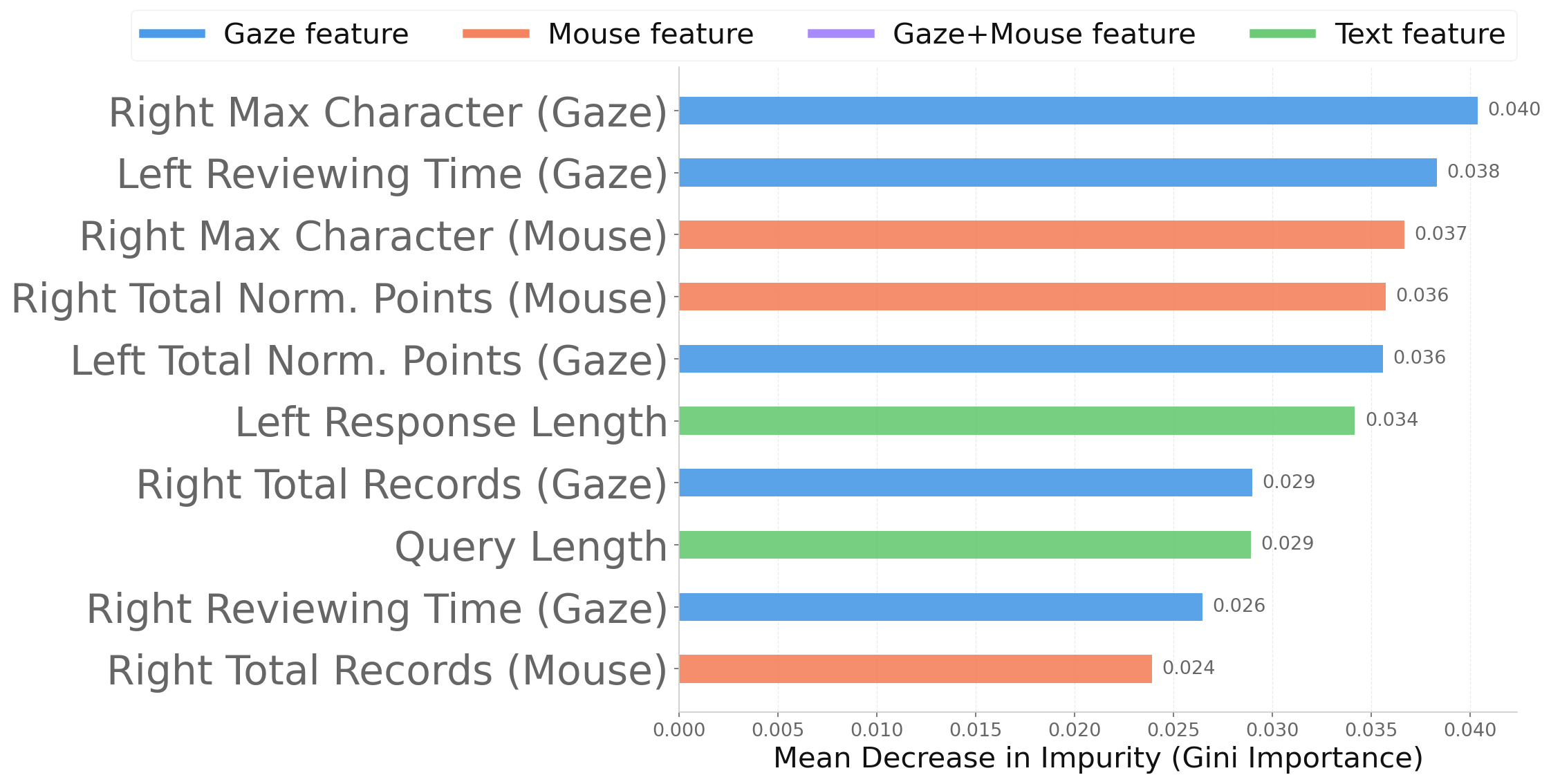}
  \captionof{figure}{The importance weights of the top 10 features for our random forest model}
  \label{fig:feature_imp}
\end{minipage}
\end{figure*}



\subsection{Reward Model Training and Analyses}
To generate high-quality chat data, we select widely-used LLMs to generate responses, which usually do not have obvious errors for the users who are not familiar with the topic. This makes their preferences hard to predict. \Cref{tab:reward_results} shows that the zero-shot performances of Claude Sonnet 4.6 and Gemma-4 31B are close to $0.5$, the level of random guesses. The supervised learning without implicit feedback also leads to similar performances. In \Cref{tab:reward_results}, all the other methods conduct 5-fold cross-validation on $695$ pairwise queries. For the standard reward models that take only the query and responses as the feature, the accuracy could only reach around $0.55$ regardless of the size of the reward models. 

To identify the useful features from implicit feedback, we first train a random forest (RF) on all features described in \Cref{sec:feature_extraction} of all data and keep only the top $50$ features with the highest weights
as our important feature (IF). We surprisingly find that the implicit feedback overpowers many features we considered effective, such as the identity of which LLMs generate the response and which user labels the preference. In both pairwise and pointwise settings, \textbf{RF + IF} achieves the best results.

To know the importance of gazing and mouse signal, we also train the random forest without mouse data and without gazing data  (i.e., IF - Mouse/Gaze). Compared to \textbf{RF + (IF - Gaze)}, the worse performance \textbf{RF + (IF - Mouse)} in \Cref{tab:reward_results} suggests that removing mouse data is more detrimental than removing gazing data.

To understand why the mouse feature is so effective, we train three random forests in data that only have short, medium, and long responses and compare the total weights of the features of each signal source in \Cref{fig:feature_imp_length}. The results show that random forest relies much more on the gazing data when the response is short. This suggests that some effectiveness of mouse signal comes from users' scrolling need because they might not point the mouse to the short responses they are reading. The complex interactions between the response length and implicit feedback features also justify our usage of random forest.\footnote{We also tried logistic regression which underperforms random forest.} In \Cref{fig:feature_imp_length}, we also observe that the response lengths, which are often the most important features in the standard reward models~\cite{singhal2024long}, have much smaller weights than the implicit feedback.


\Cref{fig:feature_imp} visualizes the top $10$ feature importance for \textbf{RF + IF} model. We can see that mouse and gazing both play important roles, and they are more important than the text length features, which are usually the strongest signal in the standard reward model~\cite{singhal2024long,dubois2024length}. The various types of time features are ranked high because the users tend to spend more time on the response they like. 

To analyze the feature influences on the prediction, we run the partial dependency analysis~\cite{friedman2001greedy}, which plots the preference prediction changes by only varying the value of a feature on average across every sample. \Cref{fig:partial_dep} shows that the higher \textit{Max Character}, the more likely they prefer the response because the users who like a response tend to finish reading it. However, the effect tends to saturate when the user only reads a little or has read a lot.

\section{LLM Alignment}
\Cref{sec:pref_pred} demonstrates that implicit feedback, especially mouse movement, could drastically improve the accuracy of reward models predicting human preference. The next research question we investigate in this section is whether better reward models in the pairwise setting could be translated into better LLM alignment outcomes. 

\subsection{Training}
For each reward model, we collect the predictions of 5 validation sets from the 5-fold cross-validation on the pairwise data. The 20\% of these predictions are used as validation data.
Our experiments test eight 1-4B base models, including GPT2-XL~\citep{radford2019language}, Pythia 2.8B~\citep{biderman2023pythia}, OLMo2 1B~\cite{walsh20252}, Llama3.2 3B~\cite{grattafiori2024llama}, Qwen2.5 1.5B, Qwen2.5 3b~\cite{hui2024qwen2}, Qwen3 1.7b, and Qwen3 4B~\cite{yang2025qwen3}. Each of the base LLM is first supervisedly fine-tuned (SFT) on all LLMs' responses.
Next, we choose to use DPO that maximizes the probability of chosen data while minimizing the rejected response probability instead of conducting reinforcement learning~\cite{ouyang2022training}, which is more expensive and often sensitive to hyperparameters, To further stabilize our experiments and avoid overfitting, we use rDPO~\cite{chowdhury2024provably} with negative log likelihood (NLL) \cite{pang2024iterative} for one epoch. Compared to the standard DPO, rDPO could emphasize the responses that are very confidently chosen by the reward model and NLL means adding an SFT loss on the chosen response. The \textbf{Explicit Feedback} from workers (i.e., the preference annotation) does not have a confidence, so we use DPO + NLL instead.

\begin{table}[t]
\centering
\scalebox{0.65}{
\begin{tabular}{lccc}
\toprule
\textbf{Reward Source} & \textbf{DPO/SFT Win Rate} & \textbf{DPO - SFT} & \textbf{Length} \\
\midrule
\vspace{-0.15cm}
Explicit Feedback  & 0.4950 / 0.4625 & 0.1079 & 	208.8 \\
& $_{\pm0.0127}$ / $_{\pm0.0161}$ & $_{\pm0.0529}$ & $_{\pm2.5}$ \\
\midrule
\vspace{-0.15cm}
mBERT base + Text  & 0.4942 / 0.4554  & 0.1221 & 227.3 \\
& $_{\pm0.0086}$ / $_{\pm0.0088}$ & $_{\pm0.0542}$ & $_{\pm2.5}$ \\
\vspace{-0.15cm}
mBERT base + Text + IF  &  0.5071 / 0.4500  & 0.1958 & 221.1 \\
& $_{\pm0.0073}$ / $_{\pm0.0082}$ & $_{\pm0.0538}$ & $_{\pm2.5}$ \\
\midrule
\vspace{-0.15cm}
RF + (IF - Gaze) & \textbf{0.5354} / \textbf{0.4196} & \textbf{0.3450} & 202.5 \\
& $_{\pm0.0119}$ / $_{\pm0.0113}$ & $_{\pm0.0531}$ & $_{\pm2.6}$ \\
\vspace{-0.15cm}
RF + (IF - Mouse) & 0.4871 / 0.4667 & 0.1108 & 203.5\\
& $_{\pm0.0113}$ / $_{\pm0.0107}$ & $_{\pm0.0541}$ & $_{\pm2.5}$ \\
\vspace{-0.15cm}
RF + IF & 0.5008 / 0.4592 & 0.1892 & 206.6 \\
& $_{\pm0.0139}$ / $_{\pm0.0131}$ & $_{\pm0.0538}$ & $_{\pm2.5}$ \\
\bottomrule
\end{tabular}
}
\caption{Average response quality of 8 LLMs after DPO using different reward models. The quality is judged by GPT4.1-mini and averaged across $2400$ prompts. Higher DPO winning rate and lower SFT winning rate is better. DPO - SFT means their average overall score difference. Explicit Feedback uses DPO + NLL, while other methods use rDPO + NLL. The standard errors are provided as our confidence region.}
\label{tab:DPO}
\end{table}

\subsection{Testing}
We randomly choose $300$ pointwise queries for testing. GPT4.1-mini compares the responses from LLM after SFT and from LLM after (r)DPO + NLL and outputs the overall scores for each response from $1$ to $10$. They are tied for the same score. Otherwise, label DPO or SFT wins. In \Cref{sec:alignment_human}, we also conduct human experiment to validate the scores from the LLM as a judge.

The results in \Cref{tab:DPO} show that the preference predictions from \textbf{mBERT base + Text} only slightly improve the output response quality after rDPO + NLL, while \textbf{RF + (IF - Gaze)} achieves much better results with average $202.5$ response length, which is significantly shorter than $228.6$, the average length of SFT responses. \Cref{tab:DPO_all} shows that the improvement is especially obvious for the recent models such as Qwen series.


The similar performances of \textbf{Explicit Feedback}, \textbf{mBERT base + Text}, and \textbf{RF + (IF - Mouse)} highlight the importance of the mouse movement signal. The unsatisfactory performances of \textbf{Explicit Feedback} show the importance of leveraging the confidence in rDPO. We also report the average performances of short, medium, and long response separately in \Cref{tab:DPO_length}. The improvement gap of \textbf{RF + (IF - Gaze)} steadily increases as the length of responses increases, while \textbf{RF + IF} seems to overfit the gazing data noise and degrade LLMs' capability for generating longer responses.










\section{Conclusion}

We introduced \dataname, a dataset pairing webcam-based eye-gaze trajectories and mouse movements with explicit preference annotations. 
The dataset allows us to systematically measure the value of implicit feedback from users for the first time. The users exhibit complicated reading patterns, which are influenced by response length, interface layout, and individual style. 
Driven by the scrolling need for the long responses, users' mouse movement trajectories carry strong preference signal that text or even eye-gazing data cannot capture and drastically improve the accuracy of reward models and response quality from the resulting aligned LLMs. The effectiveness and accessibility of the mouse movement suggest a natural path toward a self-reinforcing data flywheel driven by ordinary user interactions.









\section*{Ethical Considerations}
To collect eye-gazing and mouse trajectories data, we follow the protocol in our institution to acquire the IRB approval. We do not record video and all MTurk worker ID are anonymized before we release the data. 

Our research might bring some positive impacts such as improving the factuality evaluation~\cite{wanner2025all} by emphasizing the parts the users might pay more attention to. In contrast, our research might encourage more companies to track user's mouse trajectories or even eye movements without users' consent, which might infringe users' privacy. Besides, data flywheel might reduce the diversity of possible LLM choices in the future.





\section*{Limitations}
One limitation is that our reward model requires the implicit feedback as the input, which means at each round of RLHF~\cite{ouyang2022training}, we need to show the responses generated by LLMs to the users to collect the required implicit feedback.

Due to the page limit, we haven't analyzed some the data we collected such as the likert score for each response and answers for post-QA questionnaires. To simplify our experiments, we also split the multi-turn question-answering into multiple single-turn question-answering sessions and leave the usage of cross-session context and signals as our future work. 

\section*{Acknowledgement}
This work was supported in part by the Center for Intelligent Information Retrieval, in part by the Office of Naval Research contract \#N000142412612, and in part by Cisco. Any opinions, findings and conclusions or recommendations expressed in this material are those of the authors and do not necessarily reflect those of the sponsor.










\bibliography{custom}

\clearpage

\appendix


\section{More results}

\begin{figure}[t!]
  \includegraphics[width=0.99\linewidth]{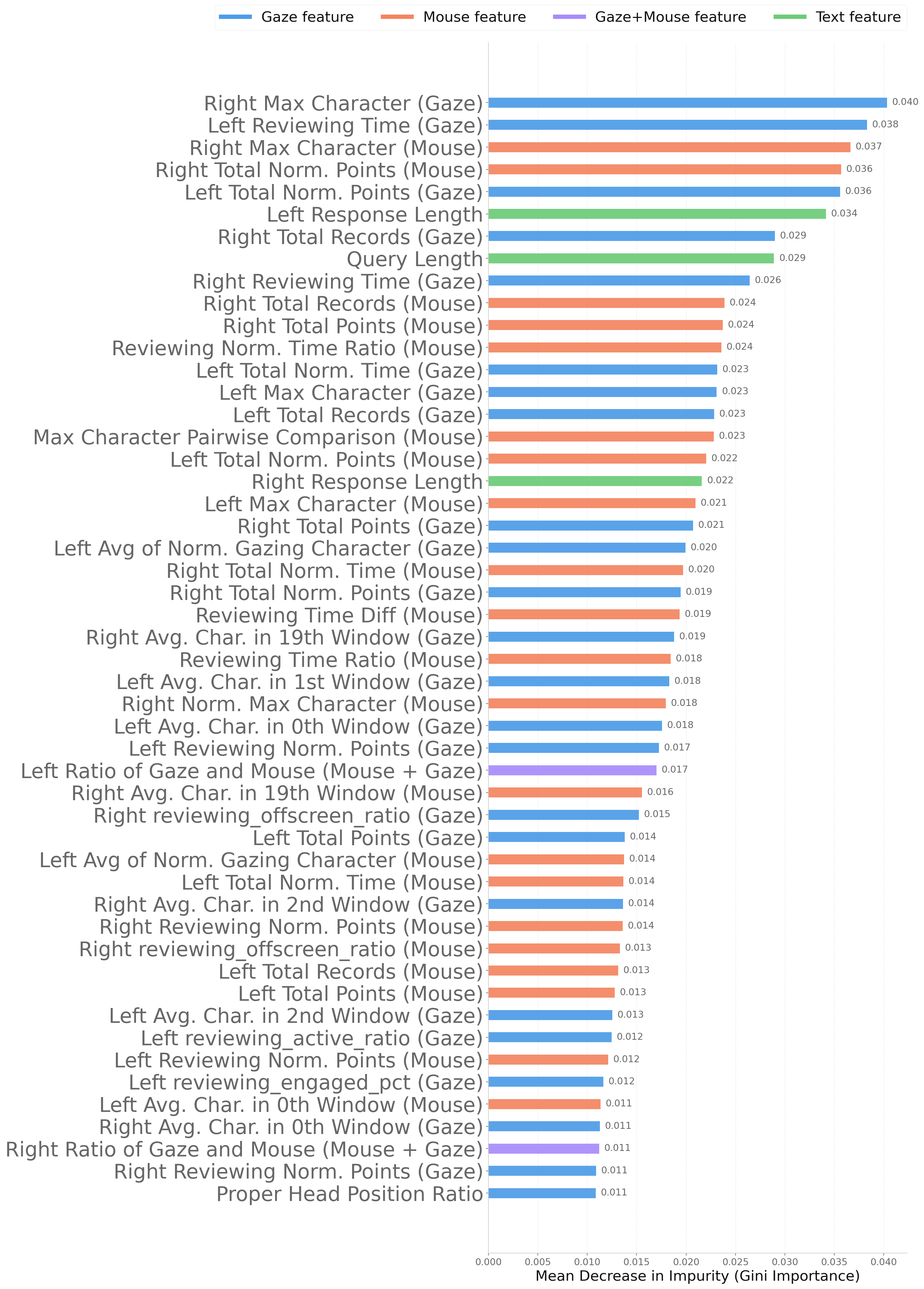} 
  \caption {The importance weights of the top 50 features for our random forest model}
  \label{fig:feature_top50}
\end{figure}

\begin{figure}[t!]
  \includegraphics[width=0.99\linewidth]{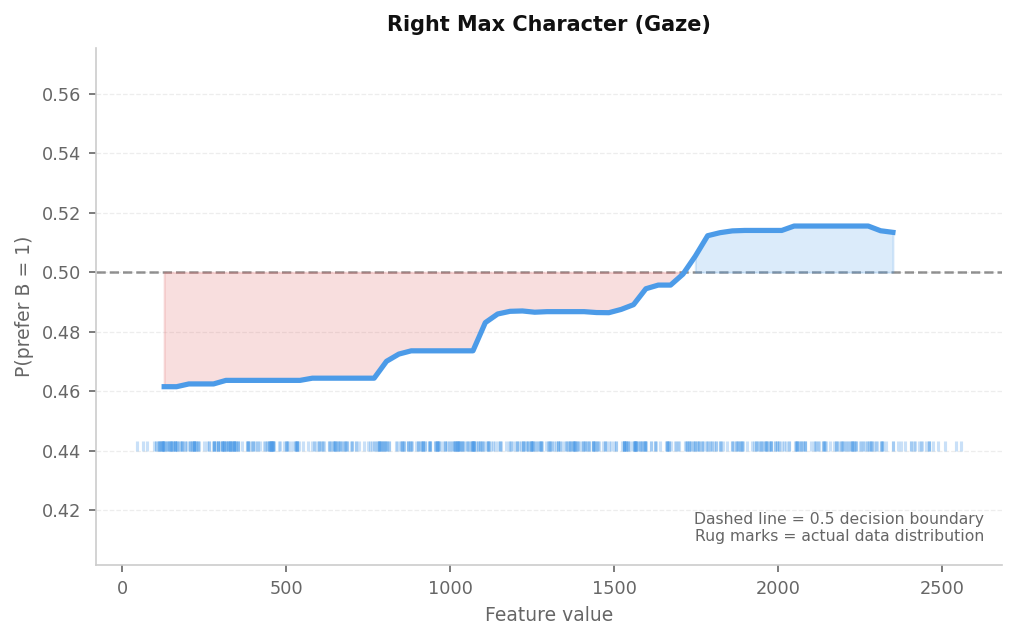} 
  \caption {Partial dependency analysis on the last character index the user gazes at the right response. As the user reads right response further, the likelihood of the right response preference increases.}
  \label{fig:partial_dep}
\end{figure}

\subsection{Preference Prediction}

In \Cref{fig:feature_top50}, we show the importance weights of all 50 features. We can see that most \textbf{Average of Characters in a Time Window} is pruned except the beginning and the end.

\begin{table}[t]
\centering
\scalebox{0.6}{
\begin{tabular}{l|cc|cc}
\toprule
 & \multicolumn{2}{c|}{Overall Human} & \multicolumn{2}{c}{Overall LLM}  \\
\midrule
\textbf{Reward Source} & \textbf{DPO-SFT} &  \textbf{H vs H} & \textbf{DPO-SFT} & \textbf{H vs LLM} \\
mBERT base + Text  & -0.407 & 0.351 &  0.470 & 0.355  \\
RF + IF &  0.383 & 0.252 & 0.470 & 0.495	 \\
\midrule
& \multicolumn{2}{c|}{Comparison Human} & \multicolumn{2}{c}{Comparison LLM} \\
\textbf{Reward Source} & \textbf{DPO/SFT} &  \textbf{H vs H} & \textbf{DPO/SFT} & \textbf{H vs LLM} \\
\midrule
mBERT base + Text  & 0.288 / 0.475 & 0.388 & 0.400 / 0.533 & 0.233 \\
RF + IF & 0.500 / 0.367 & 0.352 & 0.567 / 0.400  & 0.277 \\
\bottomrule
\end{tabular}
}
\caption{Comparison of human annotation and LLM as a judge. The responses from LLama3.2 3B for the first 30 prompts are judged. H vs H and H vs LLM mean Spearman correlation coefficient between the annotations from one MTurk worker to another worker or to LLM, respectively.}
\label{tab:DPO_human}
\end{table}

\begin{table*}[t]
\scalebox{0.63}{
\begin{tabular}{l|cc|cc|cc}
\toprule
                       & \multicolumn{2}{c|}{Short}            & \multicolumn{2}{c|}{Medium}       & \multicolumn{2}{c}{Long}         \\
                       \textbf{Reward Source} & \textbf{DPO/SFT Score} & \textbf{DPO/SFT Win Rate}     & \textbf{DPO/SFT Score} & \textbf{DPO/SFT Win Rate} & \textbf{DPO/SFT Score} & \textbf{DPO/SFT Win Rate} \\
\midrule
Explicit Feedback      & \textbf{4.513}/4.385   & 0.494/0.456          & 4.944/4.909   & 0.485/0.476      & 5.456/\textbf{5.295}   & 0.506/0.455      \\
ModernBERT + Text      & 4.447/4.379   & 0.469/0.474          & 4.856/4.830   & 0.481/0.468      & 5.621/5.350   & 0.532/0.425      \\
ModernBERT + Text + IF & 4.594/4.346   & \textbf{0.520}/\textbf{0.439} & 5.001/4.921   & 0.477/0.476      & 5.599/5.339   & 0.524/0.435      \\
RF + (IF - Gaze)       & 4.635/4.417   & 0.481/0.469          & \textbf{5.104}/\textbf{4.719}   & \textbf{0.549}/\textbf{0.405}      & \textbf{5.820}/5.388   & \textbf{0.576}/\textbf{0.385}      \\
RF + (IF - Mouse)      & 4.492/4.420   & 0.475/0.481          & 4.939/4.947   & 0.466/0.486      & 5.577/5.309   & 0.520/0.432      \\
RF + IF                & 4.484/\textbf{4.280}   & 0.501/0.463          & 5.013/4.880   & 0.486/0.477      & 5.599/5.367   & 0.515/0.438   \\
\bottomrule
\end{tabular}
}
\caption{Average quality of the responses with different lengths for 8 LLMs after DPO using different reward models.}
\label{tab:DPO_length}
\end{table*}

\begin{table*}[t]
\centering
\scalebox{0.65}{
\begin{tabular}{l|cc|cc|cc|cc}
\toprule
 &  \multicolumn{2}{c|}{GPT2-XL}  & \multicolumn{2}{c|}{Pythia 2.8B}  & \multicolumn{2}{c|}{OLMo2 1B}  & \multicolumn{2}{c}{Llama3.2 3B}   \\
Reward Source & Overall & Win rate & Overall & Win rate & Overall & Win rate & Overall & Win rate  \\
 \midrule
Explicit Feedback & 3.09 / 3.01  & 0.483 / 0.433 & 3.92 / \textbf{3.69} & \textbf{0.520} / 0.430 & 4.44 / 4.62 & 0.440 / 0.530 & 6.20 / 6.07 & 0.510 / 0.460    \\
ModernBERT + Text & \textbf{3.21 / 2.88} & 0.487 / 0.440 & 3.93 / 3.85 & 0.513 / \textbf{0.427} & 4.42 / 4.36 & 0.487 / 0.477 & \textbf{6.48} / \textbf{5.94} & \textbf{0.543} / \textbf{0.430} \\
ModernBERT + Text + IF & 3.16 / 2.90 & 0.513 / 0.420 & 3.98 / 3.91 & 0.497 / 0.453 & 4.59 / 4.49 & 0.503 / 0.473 & 6.39 / 6.18 & 0.533 / 0.450\\
RF + (IF - Gaze) & 3.08 / 2.91 & \textbf{0.540} / \textbf{0.393} & \textbf{4.05} / 3.88 & 0.500 / 0.447 & \textbf{4.91} / \textbf{4.23} & \textbf{0.593} / \textbf{0.373} & 6.43 / 6.17 & 0.507 / 0.467 \\
RF + (IF - Mouse) & 3.04 / 3.05 & 0.467 / 0.453 & 3.73 / 3.73 & 0.467 / 0.470 & 4.79 / 4.50 & 0.497 / 0.457 & 6.20 / 6.12 & 0.517 / 0.477 \\
RF + IF & 3.02 / 3.08 & 0.463 / 0.480 & 3.91 / 3.90 & 0.453 / 0.513 & 4.56 / 4.31 & 0.530 / 0.440 & 6.37 / 5.99 & 0.527 / 0.460 \\
\midrule
 & \multicolumn{2}{c|}{Qwen2.5 1.5B}  & \multicolumn{2}{c|}{Qwen2.5 3B}  & \multicolumn{2}{c|}{Qwen3 1.7B}  & \multicolumn{2}{c}{Qwen3 4B} \\
Reward Source & Overall & Win rate & Overall & Win rate & Overall & Win rate & Overall & Win rate \\
\midrule
Explicit Feedback & 4.99 / 4.60 & 0.547 / 0.403 & 5.97 / 5.85 & 0.497 / 0.470 & 4.84 / \textbf{4.66} & 0.513 / 0.443 & 6.31 / 6.39 & 0.447 / 0.530 \\
ModernBERT + Text & 4.84 / 4.65 & 0.497 / 0.447 & 5.72 / 5.91 & 0.467 / 0.493 & 4.79 / 4.79 & 0.473 / 0.450 & 6.41 / 6.45 & 0.487 / 0.480 \\
ModernBERT + Text + IF & 5.00 / 4.61 & 0.530 / 0.430 & 6.04 / 5.76 & 0.527 / 0.433 & 4.90 / 4.82 & 0.467 / 0.477 & 6.45 / \textbf{6.26} & 0.487 / 0.463 \\
RF + (IF - Gaze) & \textbf{5.11} / 4.65 & 0.550 / 0.403 & \textbf{6.24} / 5.89 & \textbf{0.563} / \textbf{0.410} & \textbf{5.20} / 4.71 & \textbf{0.533} / \textbf{0.413} & 6.48 / 6.28 & 0.497 / 0.450 \\
RF + (IF - Mouse) & 4.90 / 4.55 & 0.533 / 0.417 & 6.10 / 5.91 & 0.513 / 0.450 & 4.96 / 4.85 & 0.453 / 0.493 & 6.30 / 6.44 & 0.450 / 0.517 \\
RF + IF & 4.95 / \textbf{4.45} & \textbf{0.560} / \textbf{0.400} & 6.01 / \textbf{5.81} & 0.510 / 0.440 & 4.91 / 4.93 & 0.457 / 0.500 & \textbf{6.51} / 6.28 & \textbf{0.507} / \textbf{0.440} \\
\bottomrule
\end{tabular}
}
\caption{Response quality of each LLM after DPO using different reward models. The quality is averaged across $300$ prompts. The maximal DPO and minimal SFT are highlighted. }
\label{tab:DPO_all}
\end{table*}

\begin{table*}[t]
\centering
\scalebox{0.7}{
\begin{tabular}{l|cccc|cccc}
\toprule
& \multicolumn{4}{c}{RF + IF} & \multicolumn{4}{|c}{mBERT base + Text} \\ 
\textbf{Metrics} & \textbf{DPO/SFT Win Rate} & \textbf{DPO-SFT} &  \textbf{H vs H} & \textbf{H vs LLM} &  \textbf{DPO/SFT Win Rate} & \textbf{DPO-SFT} &  \textbf{H vs H} & \textbf{H vs LLM} \\
\midrule
Factual & 0.283 / 0.167 &	0.183 & 0.330 & - & 0.237 / 0.305 & -0.153 & 0.148 & - \\
Informativeness & 0.283 / 0.333 & -0.183 & 0.115 & - & 0.237 / 0.254 & 0.051 & 0.164 & - \\
Relevancy & 0.317 / 0.150 & 0.267* & 0.301 & - & 0.203 / 0.220 & -0.034 & 0.221 & - \\
Overall  & 0.367 / 0.150  & 0.383* &  0.252 & 0.495    &     0.254 / 0.322 & -0.051 & 0.351 & 0.355  \\
Comparison &  0.500 / 0.367 & 0.333 & 0.352 & 0.277     &   0.288 / 0.475 & -0.407 & 0.388 & 0.233	 \\
\bottomrule
\end{tabular}
}
\caption{Comparison of human annotation and LLM as a judge. H vs H and H vs LLM mean Spearman correlation coefficient between the annotations from one MTurk worker to another worker or to LLM, respectively. * means p<0.05 }
\label{tab:DPO_human_all}
\end{table*}

\subsection{LLM Alignment}
\label{sec:alignment_human}

Although LLM as a Judge usually provides evaluations that are well correlated with human judgments~\cite{lee2024rlaif}, we spent $\$270$ on MTurk for a small scale human experiment to further verify this in our setting. To facilitate effective factuality
assessment, we asked MTurk workers to search the Internet because it’s challenging to spot trivial errors in Llama3.2 3B’s responses in the first glimpse.
Each response is annotated by two master workers.

In \Cref{tab:DPO_human}, both human and GPT4.1 mini think \textbf{RF + IF} is significantly better than \textbf{mBERT base + Text}. The long responses often make the quality judgment difficult and subjective because different annotators might like different parts of the responses. The Spearman correlations between a worker and GPT4.1 mini are similar to the inter-annotator agreement, which validates the effectiveness of our LLM-as-a-judge evaluation results. 

We present the average DPO performances given different response lengths in \Cref{tab:DPO_length}, which uses the response length from each LLM after DPO to group the responses into short, medium, and long. The results show that mouse signals are more useful for longer responses. \Cref{tab:DPO_all} shows the performances of each LLM separately. We can see that the implicit feedback boosts the performance of most LLMs more. Finally, the average performance in each human evaluation dimension are reported in \Cref{tab:DPO_human_all}, which shows that \textbf{RF + IF} make the responses more relevant and factual while being slightly less informative.

\subsection{Additional User Behavior Analyses}

\subsubsection{Example of Trajectories}

\begin{figure*}[t!]
\centering
\begin{minipage}{.47\textwidth}
  \centering
  \includegraphics[width=.7\linewidth]{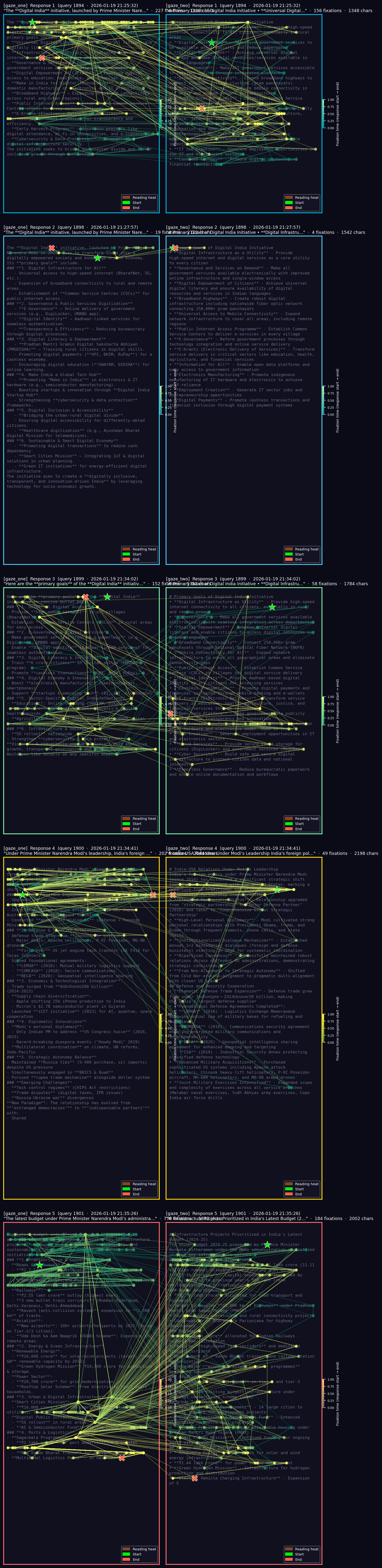}
  \captionof{figure}{An example of gazing trajectory for a topic}
  \label{fig:gaze_vis}
\end{minipage}%
\hspace{0.03\textwidth} 
\begin{minipage}{.47\textwidth}
  \centering
  \includegraphics[width=.815\linewidth]{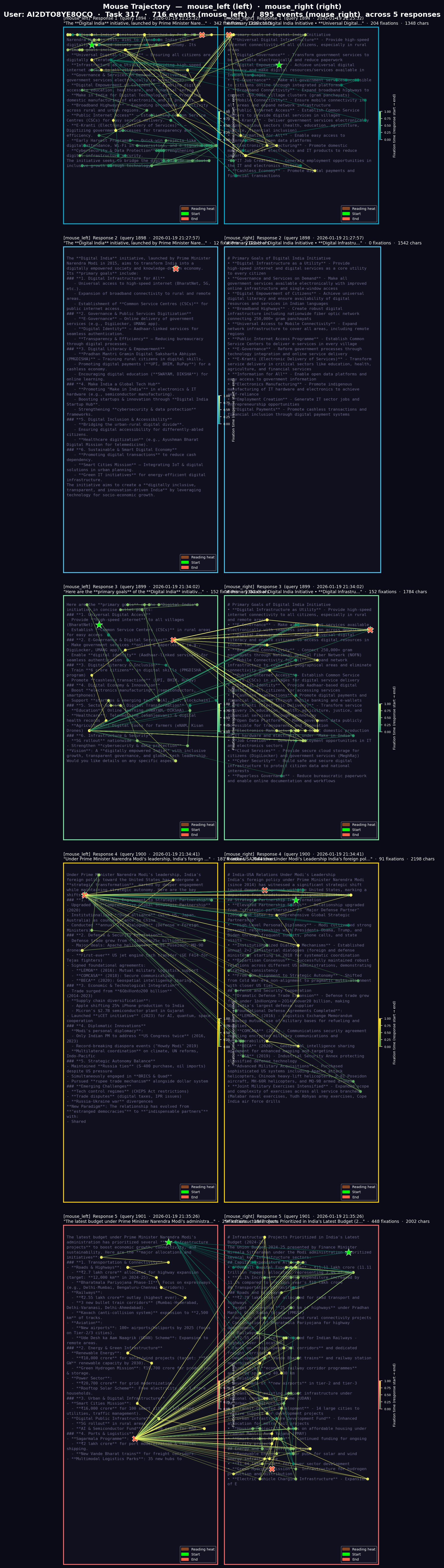}
  \captionof{figure}{An example of mouse trajectory for a topic} 
  \label{fig:mouse_vis}
\end{minipage}
\end{figure*}

We visualize the gaze trajectories from $5$-turns QA in a task from one worker in \Cref{fig:gaze_vis} and the corresponding mouse trajectories in \Cref{fig:mouse_vis}. We can see that the user could demonstrate diverse gaze behavior within a QA session.

\subsubsection{Heatmaps for Long and Short Responses}

\begin{figure*}[t!]
\centering
\begin{minipage}{.47\textwidth}
  \centering
  \includegraphics[width=.99\linewidth]{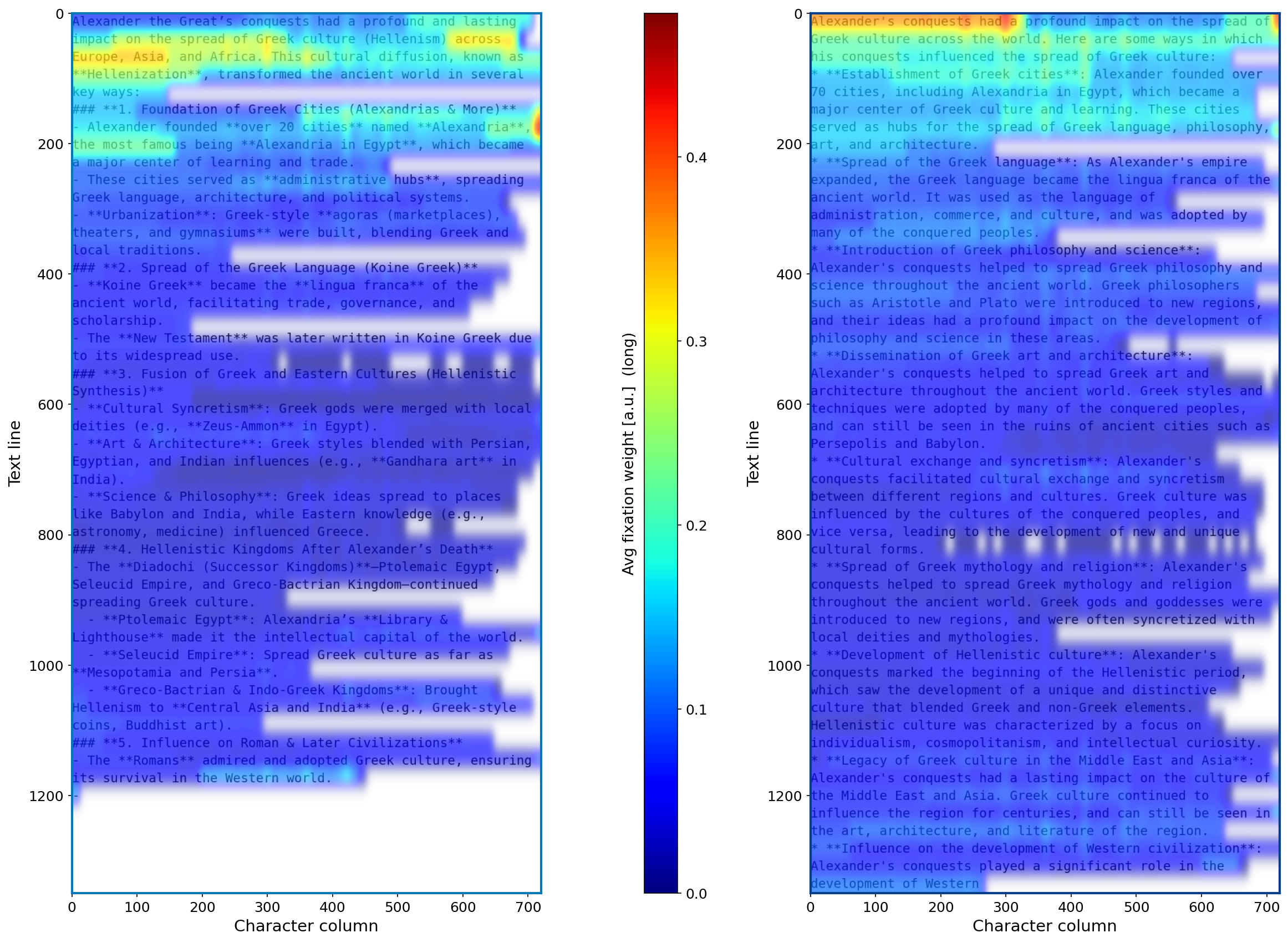}
  \captionof{figure}{Average fixation weight over the response text in the pairwise setting, aggregated across all long responses. The displayed text is a randomly selected example.}
  \label{fig:heatmap_long}
\end{minipage}%
\hspace{0.03\textwidth}
\begin{minipage}{.47\textwidth}
  \centering
  \includegraphics[width=.99\linewidth]{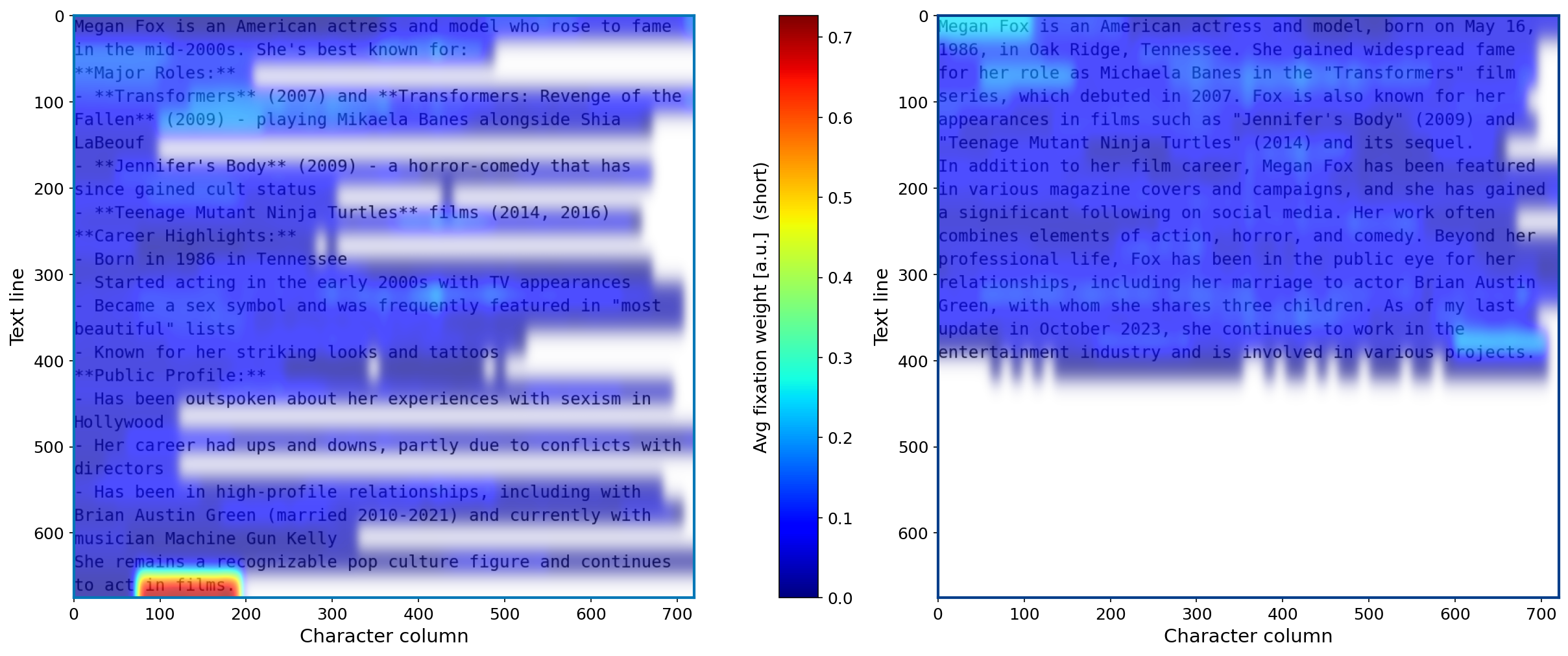}
  \captionof{figure}{Average fixation weight over the response text in the pairwise setting, aggregated across all short responses. The displayed text is a randomly selected example.}
  \label{fig:heatmap_short}
\end{minipage}
\end{figure*}

Figures~\ref{fig:heatmap_long} and~\ref{fig:heatmap_short} extend the heatmap of Figure~\ref{fig:heatmap} to long and short responses. The length-dependent pattern from Section~\ref{sec:aggregate_pattern} holds: attention stays concentrated on the early portion of long responses, while concentration weakens as responses get shorter.

\subsubsection{Mouse Reading Trajectories}

\begin{figure}[t!]
\centering
\includegraphics[width=.99\linewidth]{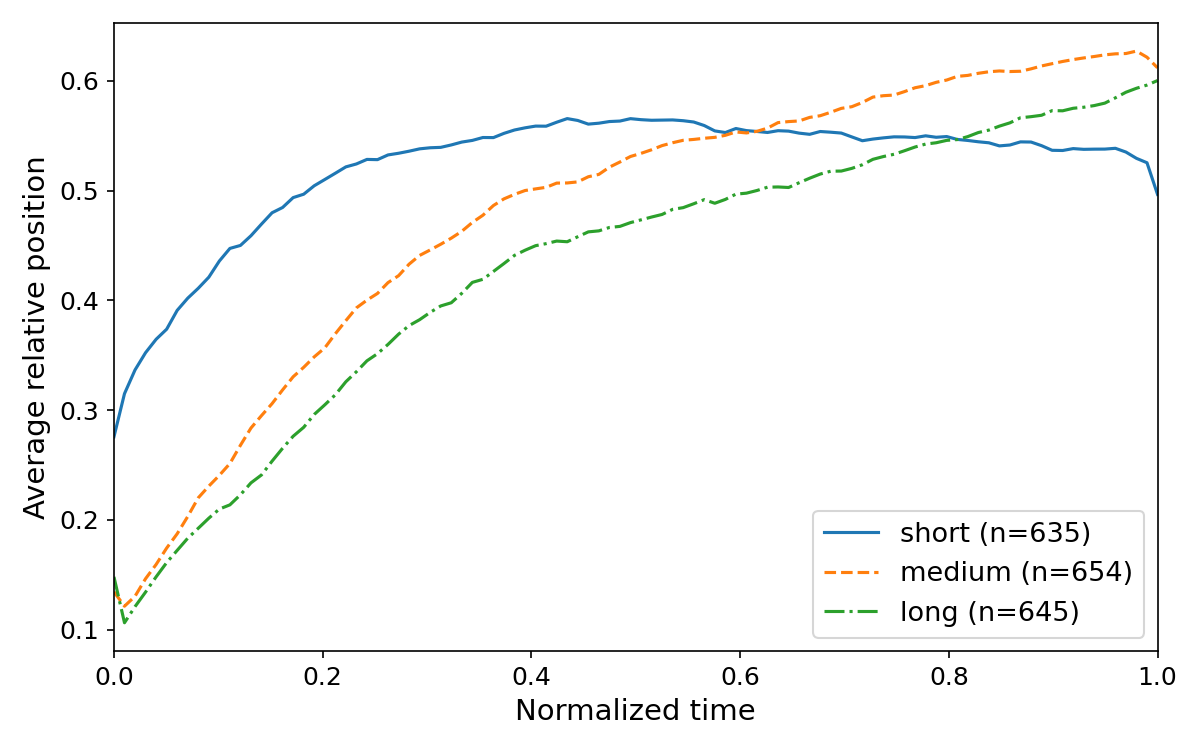}
\caption{Average mouse position over normalized time, grouped by response length.}
\label{fig:mouse_length_category}
\end{figure}

The length effect from Section~\ref{sec:aggregate_pattern} also holds for the mouse trajectory (Figure~\ref{fig:mouse_length_category}). Grouped by response length, the mouse position over time follows the same pattern as the gaze trajectory in Figure~\ref{fig:length_category}: short responses are traversed quickly and then revisited, while for longer responses the mouse stays on the early portion and advances more gradually.

\begin{figure}[t!]
\centering
\includegraphics[width=.99\linewidth]{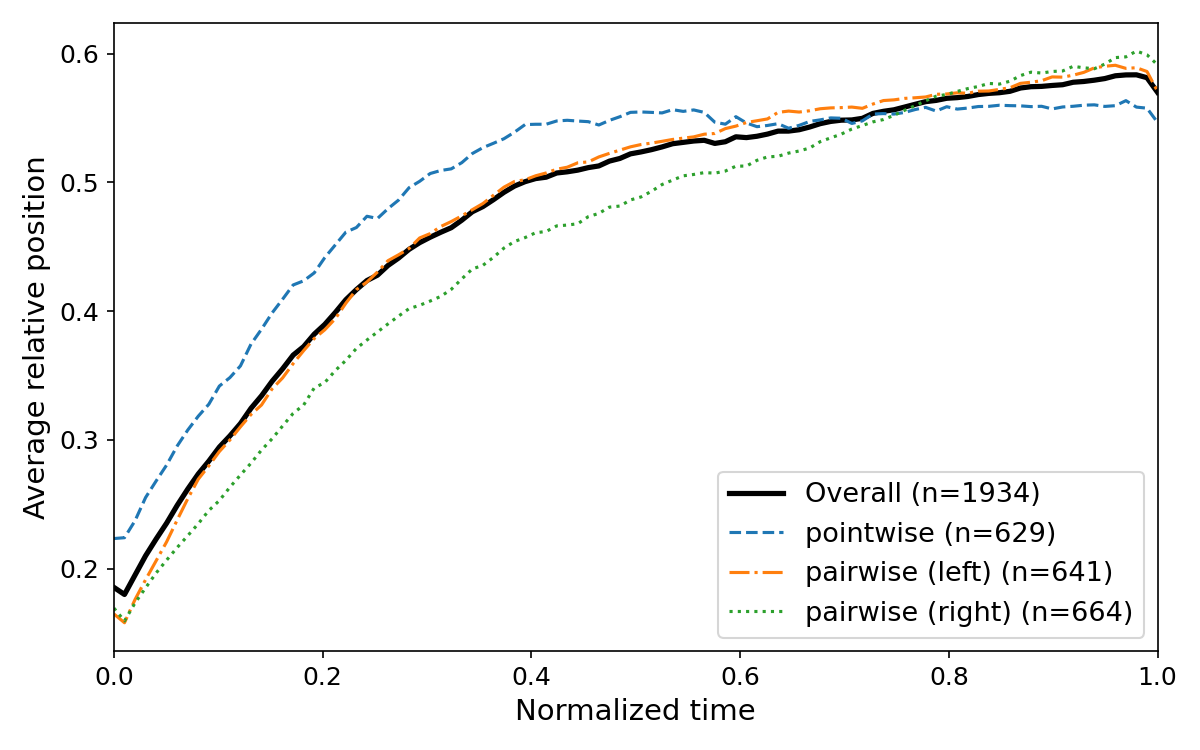}
\caption{Average mouse position over normalized time, for the pointwise setting and for the left and right responses in the pairwise setting.}
\label{fig:mouse_position_setting}
\end{figure}

Grouped by task setting (Figure~\ref{fig:mouse_position_setting}), the mouse trajectory follows the gaze trajectory in Figure~\ref{fig:pointwise_pairwise}: the pointwise setting advances fastest, and the left response is read faster than the right.

\subsubsection{Position Distributions}

\begin{figure*}[t!]
\centering
\begin{minipage}{.47\textwidth}
  \centering
  \includegraphics[width=.99\linewidth]{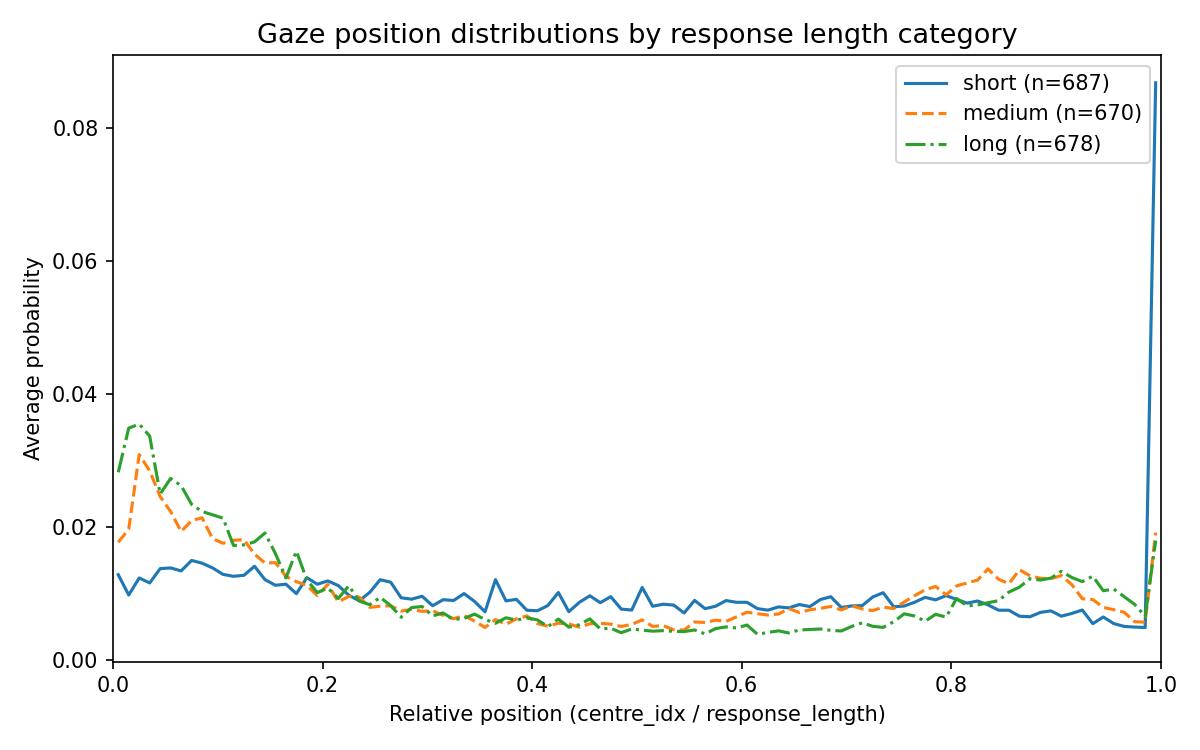}
  \captionof{figure}{Gaze position distribution across the response, grouped by response length.}
  \label{fig:gaze_length_hist}
\end{minipage}%
\hspace{0.03\textwidth}
\begin{minipage}{.47\textwidth}
  \centering
  \includegraphics[width=.99\linewidth]{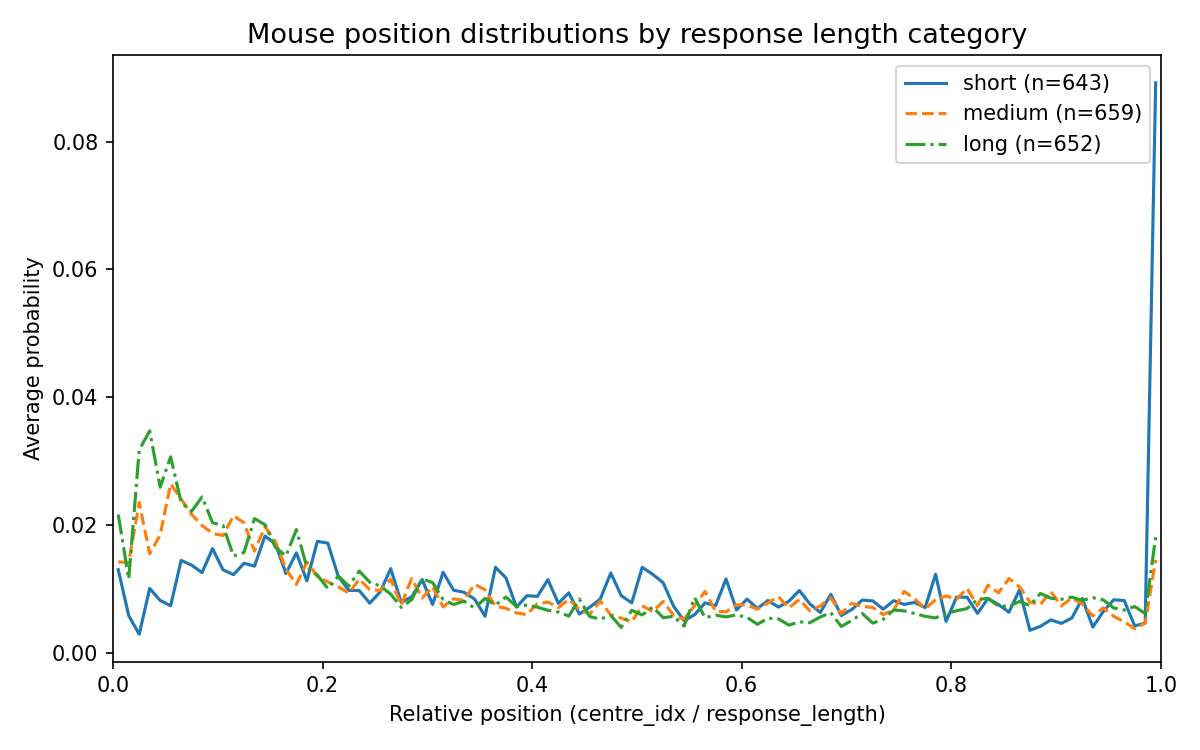}
  \captionof{figure}{Mouse position distribution across the response, grouped by response length.}
  \label{fig:mouse_length_hist}
\end{minipage}
\end{figure*}

\begin{figure*}[t!]
\centering
\begin{minipage}{.47\textwidth}
  \centering
  \includegraphics[width=.99\linewidth]{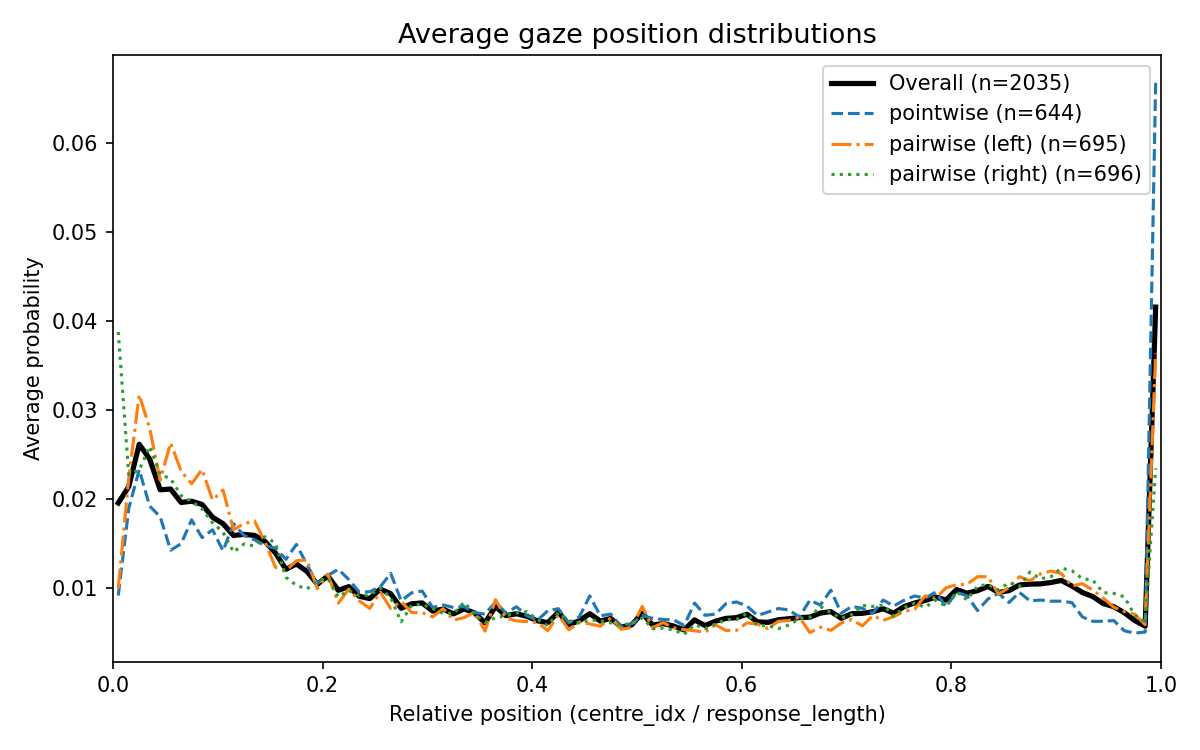}
  \captionof{figure}{Gaze position distribution across the response, for the pointwise setting and for the left and right responses in the pairwise setting.}
  \label{fig:gaze_position_hist}
\end{minipage}%
\hspace{0.03\textwidth}
\begin{minipage}{.47\textwidth}
  \centering
  \includegraphics[width=.99\linewidth]{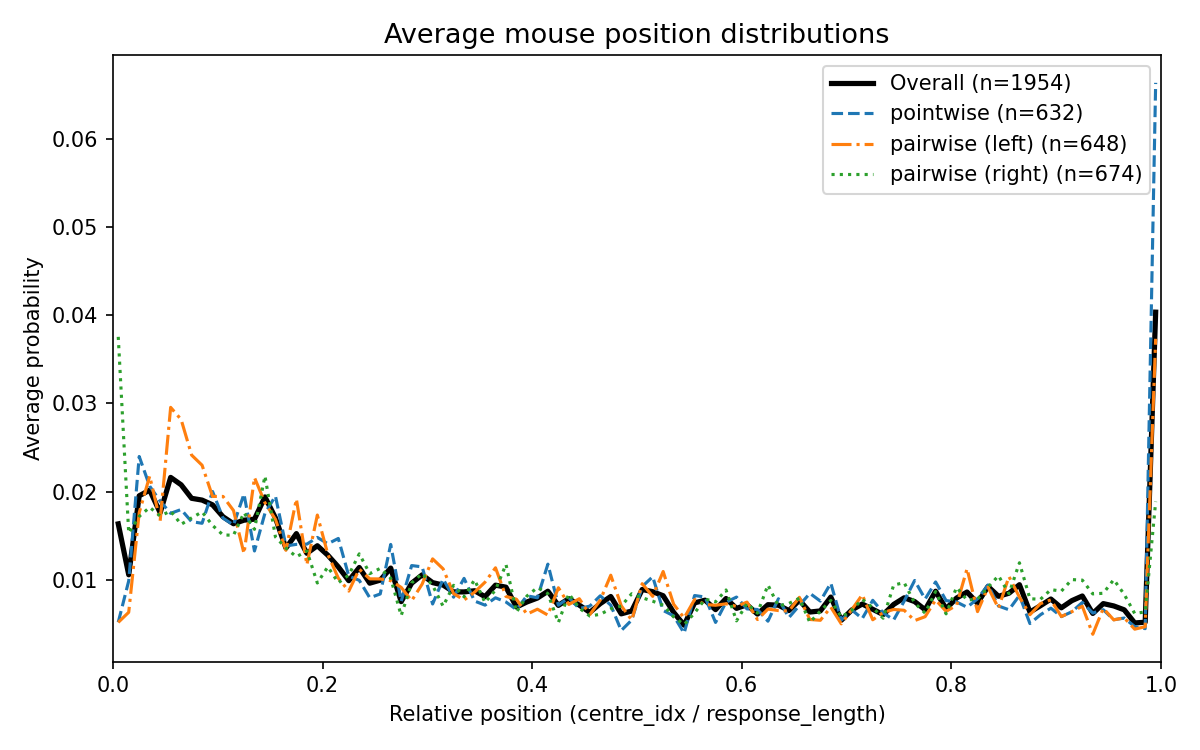}
  \captionof{figure}{Mouse position distribution across the response, for the pointwise setting and for the left and right responses in the pairwise setting.}
  \label{fig:mouse_position_hist}
\end{minipage}
\end{figure*}

The reading position can also be viewed spatially as the distribution of attention across the response. Grouped by response length (Figures~\ref{fig:gaze_length_hist} and~\ref{fig:mouse_length_hist}), attention concentrates on the early portion of medium and long responses, while for short responses a large share of it falls at the end. Grouped by task setting (Figures~\ref{fig:gaze_position_hist} and~\ref{fig:mouse_position_hist}), the distributions are largely similar across the pointwise and pairwise conditions, with attention concentrated near the start and end of the response in all three.

\subsubsection{Gaze--Mouse Correlation}

\begin{figure*}[t!]
\centering
\begin{minipage}{.47\textwidth}
  \centering
  \includegraphics[width=.99\linewidth]{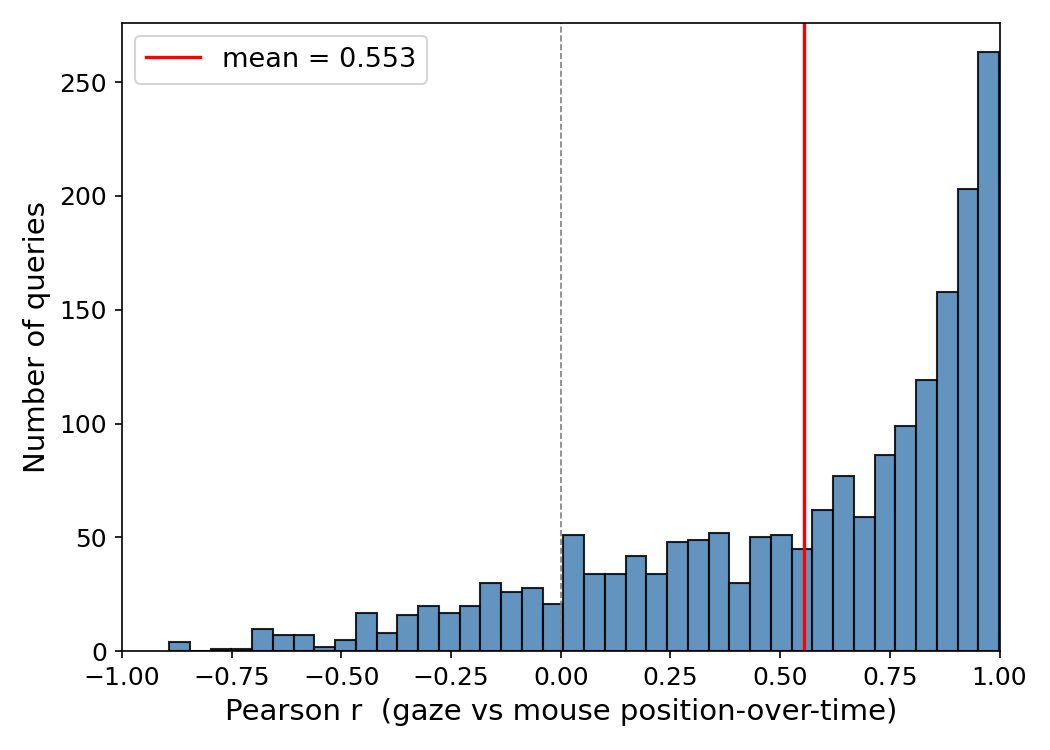}
  \captionof{figure}{Distribution of the per-query Pearson correlation between mouse and gaze position over normalized time.}
  \label{fig:corr_overall}
\end{minipage}%
\hspace{0.03\textwidth}
\begin{minipage}{.47\textwidth}
  \centering
  \includegraphics[width=.99\linewidth]{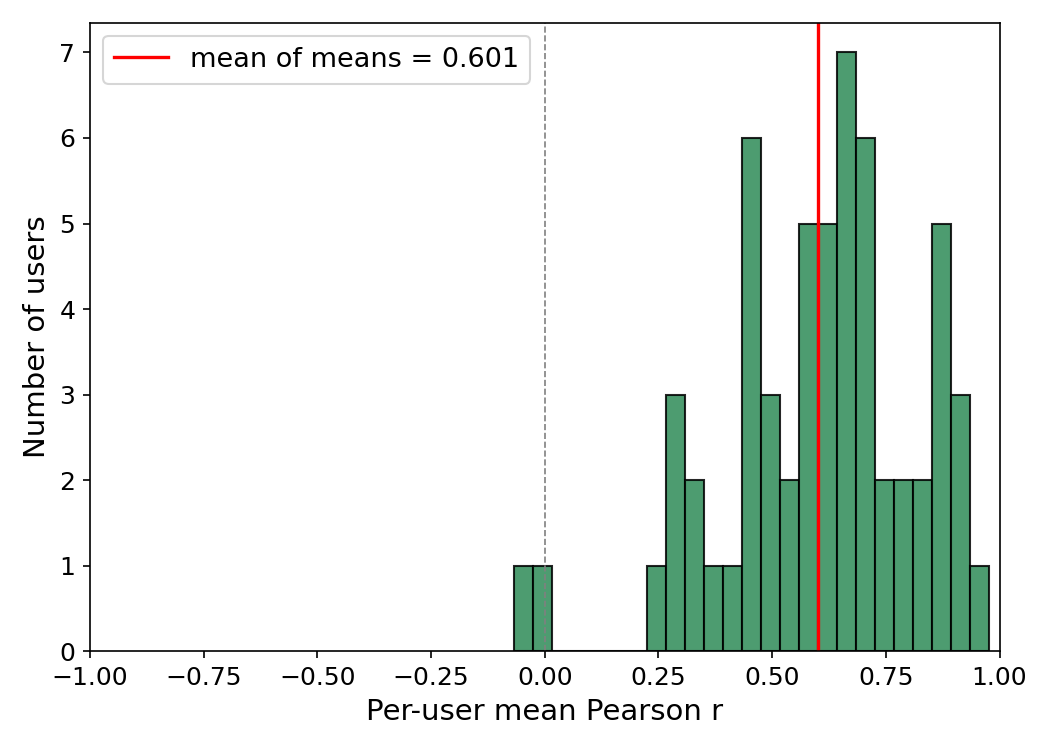}
  \captionof{figure}{Distribution of the per-user mean Pearson correlation between mouse and gaze position over normalized time.}
  \label{fig:corr_per_user}
\end{minipage}
\end{figure*}

\begin{figure}[t!]
\centering
\includegraphics[width=.99\linewidth]{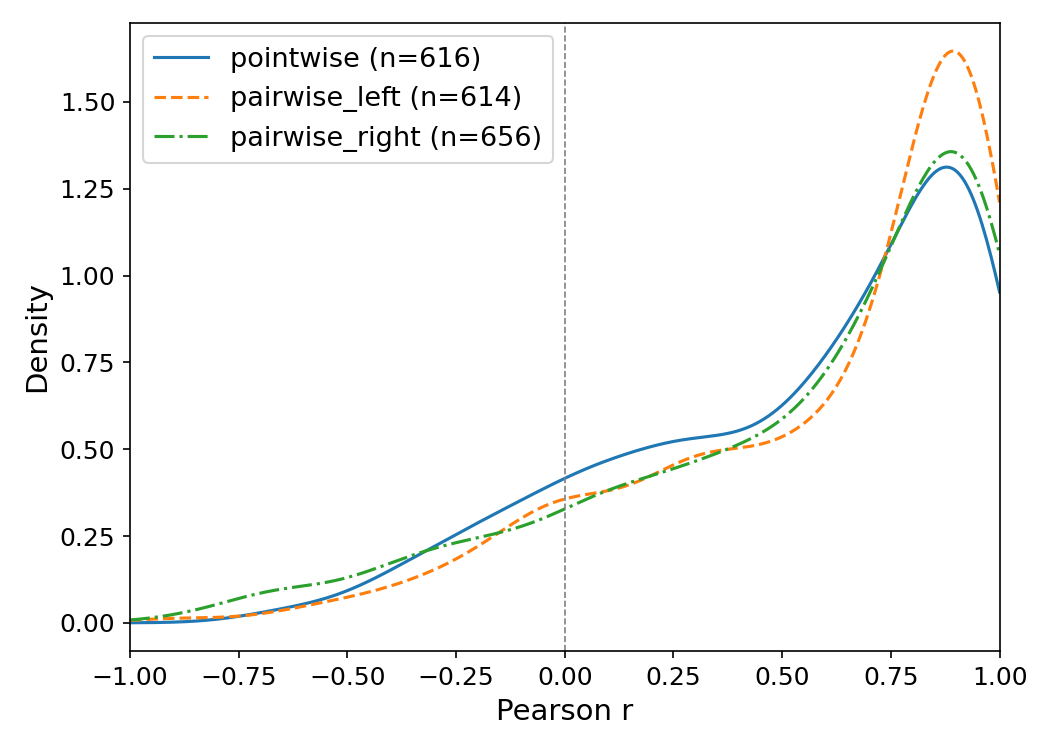}
\caption{Distribution of the per-session Pearson correlation between mouse and gaze position, for the pointwise setting and for the left and right responses in the pairwise setting.}
\label{fig:corr_by_side}
\end{figure}

The mouse and gaze trajectories are positively correlated. Across all queries (Figure~\ref{fig:corr_overall}), the per-query correlation is positive on average, and computing it per user (Figure~\ref{fig:corr_per_user}) shows that nearly every user follows this pattern. Grouped by task setting (Figure~\ref{fig:corr_by_side}), the correlation is similar for the pointwise setting and for the left and right responses in the pairwise setting.

\section{Preference Prediction Details}

\subsection{Feature Extraction}
Our gaze of an QA session are stored in a file. Since one session contains multiple queries, we need to preprocess the file to know each gaze record corresponds to which query. As mentioned before, we refer to one run as a “task” with one topic. During the task, we see that Step 5 of \Cref{fig:website_overview} with "QA and Preference Annotation" is where our task-relevant eye and mouse tracking occurs (which we will refer to as “user data”). As mentioned before, about every 0.1 seconds we track the viewed character index, a short text span which includes that index (which we will refer to as the "viewed substring"), and gaze and mouse coordinates. This data is logged on a per-user, per-task basis. To connect each row of the user data with the associated query we process the rows in time sequential order. Query IDs are positive unique identifiers for each task query provided to the users. If the user is not looking at the screen, the webcam captures this, fills the relevant user data row with placeholder values and their query ID is inherited from the last matched task query (or -2 before any matches have occurred). If the user is looking at the screen, then we apply a character-windowed approach to determine what they are looking at during a particular time-step. We check whether the viewed substring appears in the source text within 15 characters before and after the tracked character index. If matched with the experiment instruction prompt we have provided to guide users, then we assign this a query ID of -1. If still unmatched, then we iterate the relevant set of task queries (either pairwise or pointwise) for said user to match the viewed substring with the relevant task query ID. On-screen data that didn't match any of the prior conditions are provided a default query ID of 0.



When computing the ratio of two features $A$ and $B$, we use $\min(A/(0.001+B), 100)$ to prevent from having a large value for small $B$. One second smoothing means that whenever we observe a gaze point side a response textbox, we assume the user still looks at that response in the next second to reduce the noise in the gazing data. Besides reviewing features, we also apply the one second smoothing is also applied to \textit{Total Norm. Time}.

\subsection{LLM Reward Model}

We use the following prompt for Gemma 4 31B and Claude Sonnet 4.6 to get their zero-shot preference prediction.
\small
\begin{lstlisting}
You are an expert evaluator assessing the quality of two AI-generated responses to a user query.
Your task is to determine which response better answers the user's query.
Output your judgment as JSON with exactly two fields:
- "prediction": 1 if Response 1 is better, 2 if Response 2 is better
- "confidence": a float between 0.0 and 1.0 indicating how confident you are (0.5 = completely uncertain, 1.0 = completely certain)
Output only valid JSON, nothing else.

User Query:
{query}

Response 1:
{response_1}

Response 2:
{response_2}

Which response better answers the user's query? Output JSON only.
\end{lstlisting}
\normalsize
There are $3$ samples out of $695$ samples that are not able to processed by Gemma 4 31B, so we ignore them when computing the performances.
When we add the important features to the ModernBERT, we simply append every feature name and its value to the text of user query and the two responses.

\subsection{Pointwise Settings}
\label{sec:pointwise}

We also train our random forest reward model on the pointwise data, where it predicts the likert score of a single response [1-5]. We use the same text features and the implicit feedback features extracted from the mouse and gaze trajectories described in Section~\ref{sec:feature_extraction}. The R$^2$ of the model is only around $0.05$ under 5-fold cross-validation.

The score is difficult to predict because each worker might have different bias toward higher or lower scores and we also notice that workers rarely give different score in a session. To force them to express their preference, we add the question of comparing with the previous question in the pointwise setting. 

We discover that the workers have a strong bias: 70\% of annotations prefer the current response compared to the previous response. To balance the prediction classes, we subsample the data that prefer the current response.



\subsection{Hyperparameters for ModernBERT and Random forest}

For ModernBERT and Qwen3 1.7B, we set the batch size to be 1 and learning rate to be 1e-5. For pairwise, the number of epoch is $10$ and for pointwise, which has fewer samples, we set the number of epoch as $5$ to reduce overfitting.

We use the random forest implementation from Scikit-learn library \cite{pedregosa2011scikit}. When identifying the feature weights, we set max depth as $5$ to capture more complex interaction and set the number of estimators as $200$, minimal split as $10$, and minimal leaf size as $4$ to reduce overfitting. For the random forest that uses the important features, we use $5$ max depth, $100$ estimators, $5$ minimal split, and $2$ minimal leaf size.


We coarsely tune the hyperparameters of ModernBERT and random forest according to our validation scores, but we found that the performances are not sensitive to these hyperparameters.




\section{LLM Alignment Details}

We modify the DPO implementation from \url{https://github.com/eric-mitchell/direct-preference-optimization} and use their default hyperparameter $\beta=0.1$ and learning rate is $5e-7$. To reduce the memory requirement, we set batch size to be $2$. All the models are trained using NVIDIA A100 80G.


We find DPO or rDPO along often decreases the loss by reducing the probability of both chosen and rejected probabilities, but reduce the rejected responses more. Adding the NLL/SFT term solves this problem.

The prompt of LLM as a judge is listed below:
\small
\begin{lstlisting}
You are an expert evaluator assessing the quality of AI assistant responses.
You will be given a conversation prompt and two responses (A and B) from different AI models.

Evaluate each response on these criteria:
1. Instruction Following: Did the model follow all explicit and implicit instructions?
2. Informativeness: Is the response comprehensive without being verbose?
3. Factuality: Are the claims accurate? For creative prompts, judge internal consistency.
4. Clarity and Coherence: Is the response well-structured and easy to read?
5. Overall Helpfulness: Which response is more ready to use for the human?

You MUST always respond in EXACTLY this format (no extra text, no markdown, no blank response):
SCORE_A: <integer 1-10>
SCORE_B: <integer 1-10>
WINNER: <A or B or tie>
REASONING: <one concise sentence>

Study these examples carefully before evaluating:

EXAMPLE 1
## Conversation Prompt
Human: What is the capital of France?

## Response A
The capital of France is Paris. It has been the country's political and cultural centre for centuries.

## Response B
France.

SCORE_A: 9
SCORE_B: 3
WINNER: A
REASONING: Response A directly and accurately answers the question with useful context, while Response B names the country instead of its capital.
---
EXAMPLE 2
## Conversation Prompt
Human: Write a short poem about autumn.

## Response A
Leaves fall like whispered secrets,
Gold and red adorn the trees,
Crisp air carries distant echoes
Of summer's last, reluctant breeze.

## Response B
Autumn is a season. Trees lose leaves. It gets cold.

SCORE_A: 9
SCORE_B: 2
WINNER: A
REASONING: Response A fulfils the creative request with imagery and rhythm; Response B is a flat, prosaic description with no poetic quality.
---
EXAMPLE 3
## Conversation Prompt
Human: How do I reverse a list in Python?

## Response A
You can reverse a list in Python using the built-in reverse() method: my_list.reverse() modifies it in place, or use my_list[::-1] to get a new reversed list.

## Response B
Use the reverse function on the list object. It will reverse the list for you.

SCORE_A: 8
SCORE_B: 5
WINNER: A
REASONING: Response A provides two concrete, correct methods with brief code examples, while Response B is vague and offers no actionable syntax.
---
Now evaluate the following pair using the EXACT same format as the examples above.
\end{lstlisting}
\normalsize



\subsection{LLM Alignment Human Experiments}

\begin{figure}[t!]
  \includegraphics[width=0.99\linewidth]{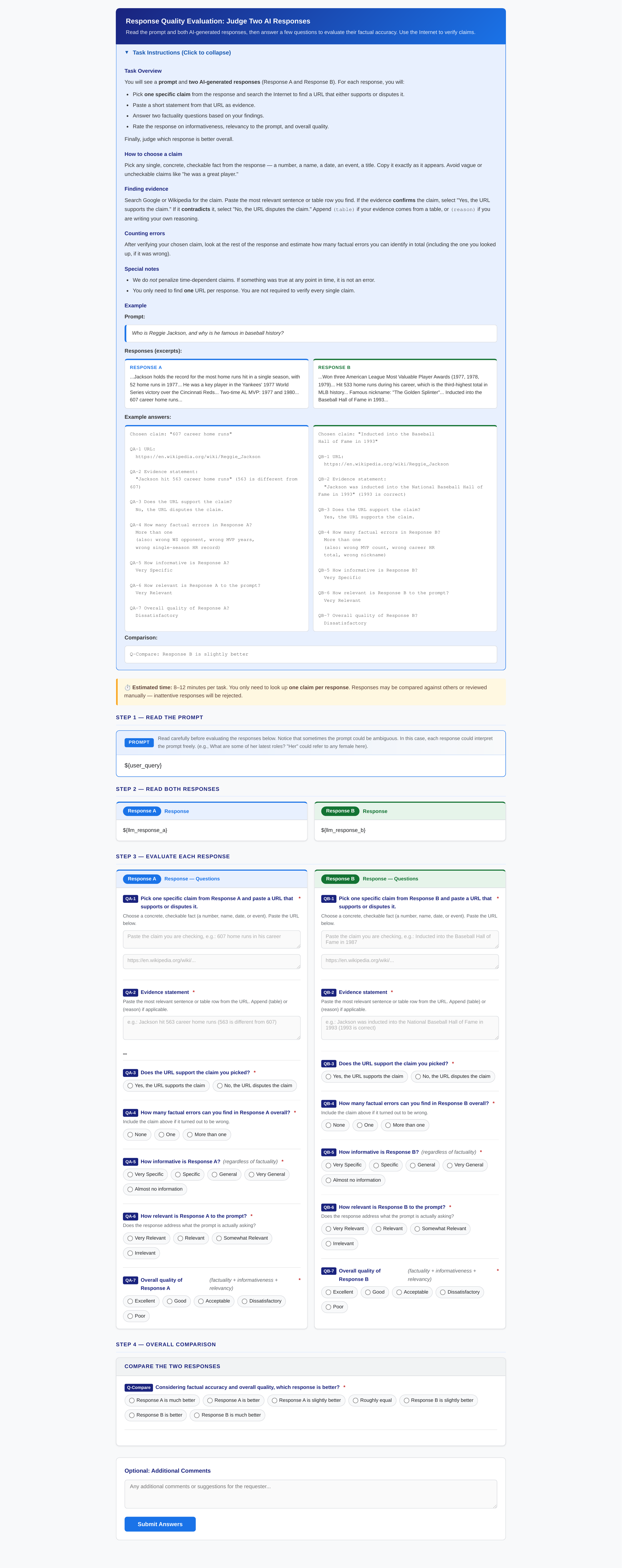} 
  \caption{The crowdsourcing template we used in our LLM alignment experiment.
  }
  \label{fig:crowd_template}
\end{figure}

Our MTurk template could be seen in \Cref{fig:crowd_template}. Only master workers could do the task. We provide \$1.6 or \$2 wage for each task, which takes around 10 minutes. We choose to test Llama3.2 3B because it could output coherent responses with some errors for workers to find. $9$ out of $120$ responses are rejected by Claude code and manual inspection.

Our website allows the users to ask follow-up questions, so some queries are ambiguous without showing the previous queries. We instruct the MTurk workers to allow the LLMs to interpret query freely (e.g., \textit{What is her most important role?} does not mention who \textit{she} refers to, so the responses from LLMs can talk about any actress).


\section{AI Usage}
We use Claude code to generate some analysis codes and MTurk Template. We also use Claude, Gemini, and ChatGPT to help us develop the website, search for some related work, or provide writing suggestions.

\section{Website Details}

Our website is developed using PHP and MySQL database. 
MTurk workers might use various types of browsers and often do multiple tasks in parallel. Each user is allowed the use of Google Chrome, Firefox, and Microsoft Edge. Multiple windows and tabs are supported but 1 tab total is encouraged. 

\subsection{Login and Pre-test Questionnaire}

We manually filtered topics for content sensitivity or being too niche for non-factoid conversation such as "Jeffrey Epstein" and "Biggest ball of twine".

The General Information Questionnaire consists of 2 pages. The first page requests for user consent of the experiment and acknowledges the use of a web camera. Note the browser itself requests for camera use as well. The second page is a questionnaire on the background of the user such as demography and highest education level with an emphasis on flexibility. We mention in the consent page all data is secured in our server. 

Both new and old users are met with the Instruction Page, Step 3. It contains a list of instructions that detail high quality queries with positive and negative examples, the webpages they can expect, and troubleshooting if the webcam does not work. 
In the instruction, we encourage workers to move the mouse to the places they gaze.
The rest of the experiment features a navigation bar with a hyperlink to the Instruction Page to further its accessibility.

Any head position out of the green box may incur poor prediction. The calibration uses 8 buttons around the screen the user must move the mouse and click multiple times. Alongside the mouse is a red dot constantly displaying the prediction of Webgazer. The red dot allows the user to understand the prediction model but remains a distraction for further steps in the experiment, hence the red dot is only for calibration. The button presses assume the user is gazing at the button with each press, acting as a ground truth for the current eye gazing point. Afterwards, the user moves the mouse and their gaze to the center to measure accuracy. If a suitable accuracy is met (refer to \ref{subsec:quality_control}, the user may proceed with the experiment.

\begin{figure}[t!]
  \includegraphics[width=0.99\linewidth]{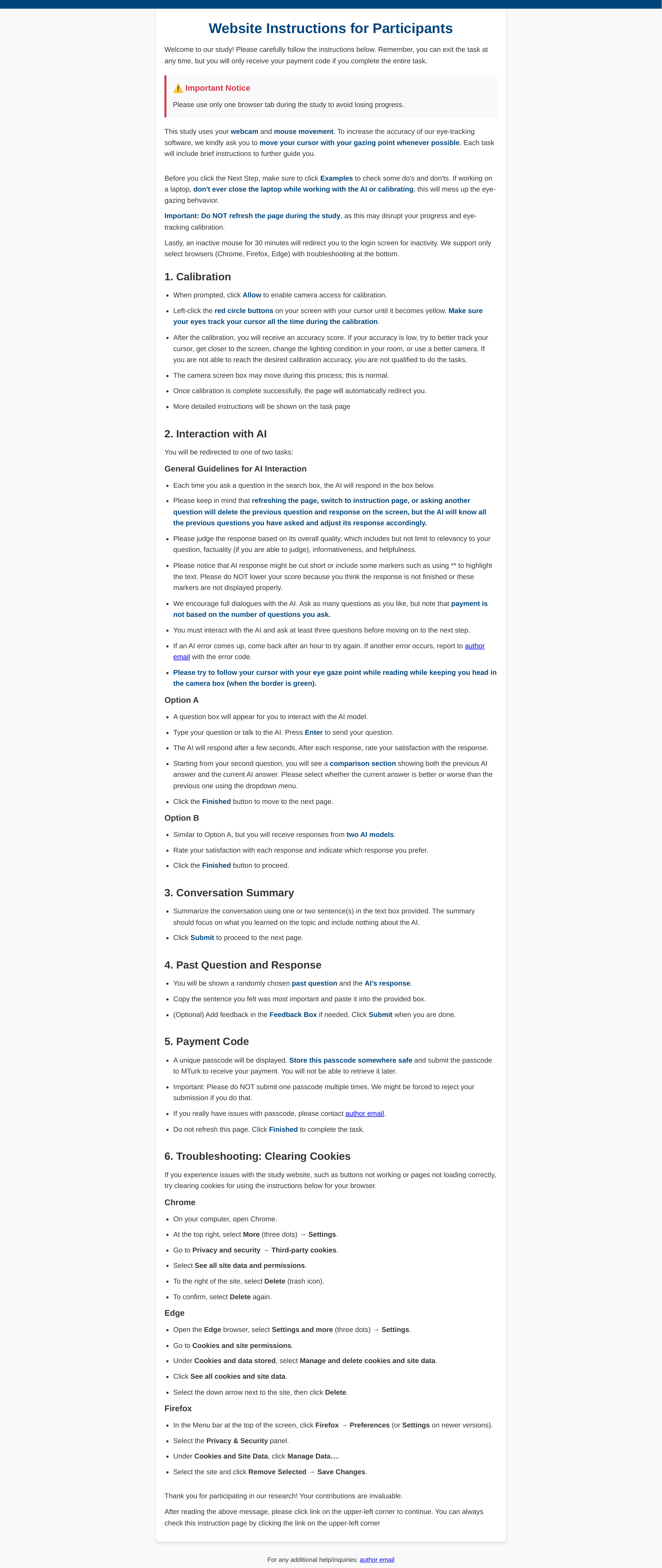} 
  \caption {Our website instruction page}
  \label{fig:instruction}
\end{figure}

\subsection{QA and Preference Annotation}

Both pointwise and pairwise contain a small instruction set at the top, a webcam, the query box, and the navigation bar. The full instruction could be seen at \Cref{fig:instruction}.

%


\subsection{Quality Control}

\begin{figure}
    \centering
    \includegraphics[width=0.99\linewidth]{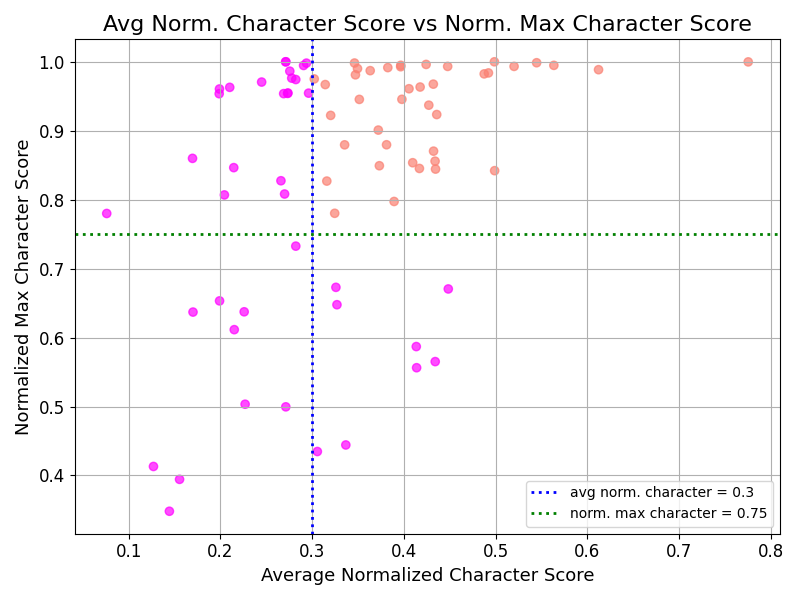}
    \caption{Macro average of each user's Average Normalized Character (i.e., Total Norm. Points) vs Norm. Max Character. Magenta points were users below thresholds, in consideration for removal of dataset}
    \label{fig:avg_norm_character_vs_norm_max_character}
\end{figure}

The variables chosen in Figure~\ref{fig:avg_norm_character_vs_norm_max_character} best represent user integrity and associated quality cutoffs (0.75 for response score and 0.3 for max index score) for the representation of the user's attention to the task.

\end{document}